\documentclass{article}
\usepackage{graphicx}
\usepackage{amsmath}

\usepackage{subcaption}

 \usepackage[dblblindworkshop, final]{neurips_2026}

\usepackage[utf8]{inputenc} % allow utf-8 input
\usepackage[T1]{fontenc}    % use 8-bit T1 fonts
\usepackage{hyperref}       % hyperlinks
\usepackage{url}            % simple URL typesetting
\usepackage{booktabs}       % professional-quality tables
\usepackage{amsfonts}       % blackboard math symbols
\usepackage{nicefrac}       % compact symbols for 1/2, etc.
\usepackage{microtype}      % microtypography
\usepackage{xcolor}         % colors
\usepackage{multirow}
\usepackage{colortbl}

\title{VisConf: Quality-Aware Consensus for Best-of-N in Vision-Language Model Reasoning}
\workshoptitle{VLM4RWD}

\author{%
  Muhammad Usman\textsuperscript{1}
  \And
  Muhammad Qasim Ayub\textsuperscript{1}
  \And
  Usman Ahad\textsuperscript{1}
  \And
  Muhammad Tahir\textsuperscript{2}
}

\begin{document}

\maketitle
\vspace{-1em}

\begin{center}
    \textsuperscript{1}Department of Computer Science \quad \textsuperscript{2}Department of Electrical Engineering\\
    Lahore University of Management Sciences\\
    \texttt{\{27100046, 27100168, 27100041, tahir\}@lums.edu.pk}
\end{center}

%%%%%%%%%%%%%%%%%%%%%%%%%%%%%%
\begin{abstract}
Best-of-N methods improve text-only LLM reasoning by sampling multiple trajectories before selecting an answer. For Vision-Language Models (VLMs), answer frequency and model confidence alone can be insufficient, as sampled generations also vary in their engagement with the visual input. We find that distributional confidence and visual attention are complementary indicators of rollout quality. Each is limited on its own, while trajectories that show strong signals in both are more likely to be correct. Motivated by this, we introduce VisConf, a rollout-quality score that multiplicatively combines visual engagement with candidate-calibrated Self-Certainty, penalizing rollouts that are weak in either signal. We incorporate Rank-Weighted Consensus (RWC), a tuning-free aggregation rule that converts within-question VisConf ranks into answer-level vote weights. Across three VLMs on MathVista, MMMU-Pro, and MMStar, VisConf with RWC consistently outperforms strong BoN baselines. Our analysis shows particularly large gains when correct trajectories are present but underrepresented in the candidate pool. \href{https://anonymous.4open.science/r/VisConf-VLM-BoN/}{\underline{Code}}.
\end{abstract}
%%%%%%%%%%%%%%%%%%%%%%%%%%%%%%

%%%%%%%%%%%%%%%%%%%%%%%%%%%%%%
\section{Introduction}
A widely used test-time scaling method in Large Language Model (LLM) reasoning is Best-of-N (BoN) selection, where multiple candidate trajectories are sampled for each query \citep{snell2024scalingllmtesttimecompute, brown2024large}. Generating multiple rollouts using stochastic decoding provides more opportunities to produce correct reasoning paths, but also introduces a selection problem since repeated sampling can produce redundant reasoning patterns rather than genuinely diverse solutions \citep{bilal2026refining}. Vision-Language Models (VLMs) similarly benefit from rollout scaling \citep{sammani2026testtimescalingvisionlanguagemodels, tong2026diversitymattersrevisitingtesttime}, but multimodal reasoning introduces another dimension of candidate variation as trajectories can differ in how strongly they engage with the visual input.

Two natural signals for assessing VLM rollout quality are confidence and visual engagement. Output-probability-based confidence captures how decisively the model supports its generated trajectory, and has been widely used as a proxy for reliability  \citep{geng2024surveyconfidenceestimationcalibration, kang2025scalablebestofnselectionlarge}. However, high confidence does not necessarily imply correct visual reasoning \citep{cho2026lookbackscorelvlmresponses}. Visual attention can indicate greater reliance on the image, but may also reflect repeated inspection caused by uncertainty or misinterpretation. Our empirical analysis shows that these signals are weakly informative but provide complementary information at the trajectory level: rollouts exhibiting both high confidence and strong visual engagement are substantially more reliable than those strong in only one dimension. We further find that their useful interaction is primarily trajectory-level rather than token-aligned. Motivated by these findings, we introduce VisConf, which independently aggregates confidence and visual engagement over a rollout before combining them into a single quality score.

For verifiable tasks, BoN exposes two sources of evidence: the quality of individual rollouts and agreement across the candidate set. Self-Consistency exploits the latter by favoring answers supported by multiple independently sampled trajectories, under the intuition that diverse reasoning paths converging to the same answer provide stronger evidence of correctness \citep{wang2023selfconsistencyimproveschainthought}. However, standard Self-Consistency treats all rollouts equally and fails when redundant incorrect generations form a majority group \citep{yin2024aggregationreasoninghierarchicalframework, taubenfeld-etal-2025-confidence}. In contrast, conventional BoN selection relies only on the single highest-scoring rollout, discarding corroborating evidence from other trajectories that share the same outcome. We therefore use Rank-Weighted Consensus (RWC), a Borda-inspired hyperparameter-free aggregation scheme, to combine trajectory quality with answer-level consensus.

Our contributions are as follows: (\(1\)) We empirically characterize confidence and visual engagement as complementary trajectory-level signals of VLM rollout quality, showing that their useful interaction is primarily trajectory-level rather than token-aligned. (\(2\)) We introduce a VLM BoN framework featuring VisConf, which combines candidate-calibrated Self-Certainty with visual engagement, and RWC, which incorporates trajectory quality into answer-level consensus. (\(3\)) We demonstrate consistent improvements across three models, three benchmarks, and three sampling budgets, including average accuracy gains of \(0.98\) points over Self-Consistency and a statistically significant \(0.49\) points over the strongest baseline. 

%%%%%%%%%%%%%%%%%%%%%%%%%%%%%%

%%%%%%%%%%%%%%%%%%%%%%%%%%%%%%
\section{Related Work}
\paragraph{Test-Time Scaling and Best-of-N}
Test-time scaling improves performance by allocating additional inference compute without modifying model parameters \citep{snell2024scalingllmtesttimecompute, brown2024large}. Self-Consistency aggregates candidates through answer frequency, but provides no explicit signal of rollout quality \citep{wang2023selfconsistencyimproveschainthought}. Alternatively, verifier- and reward-model-based methods rank candidates using learned correctness features, at the cost of training or invoking an additional model \citep{cobbe2021trainingverifierssolvemath, lightman2023letsverifystepstep}. Recent work in VLMs finds modest gains from single-model scaling while evaluating lexical/length-based heuristics and majority voting, leaving stronger model-internal rollout-quality signals unexplored \citep{tong2026diversitymattersrevisitingtesttime}.

\paragraph{Confidence-Based Selection}
Model confidence provides a verifier-free quality signal for sampled trajectories \citep{geng2024surveyconfidenceestimationcalibration, kang2025scalablebestofnselectionlarge}. Likelihood-based measures such as sequence log-probability use only the probabilities assigned to the sampled tokens, ignoring the remainder of the predictive distribution \citep{kang2025scalablebestofnselectionlarge, kim2026reasoningfluencydissectingprobabilistic}. Self-Certainty instead uses the full predictive distribution through Kullback-Leibler (KL) divergence from a uniform reference distribution \citep{kang2025scalablebestofnselectionlarge}. However, its Borda-based aggregation introduces a tunable exponent.

\paragraph{Visual Grounding in VLMs}
VLM responses can be weakly image-conditioned, exhibiting hallucination and language-prior bias, motivating methods that promote visual faithfulness \citep{li2023evaluatingobjecthallucinationlarge, Leng_2024_CVPR}. VGPO identifies temporal visual forgetting, showing that sustained visual attention can be indicative of correctness \citep{wang2026visuallyguidedpolicyoptimizationmultimodal}. Most closely related, LookBack combines token likelihood with attention-based visual reference to select the highest-scoring rollout \citep{cho2026lookbackscorelvlmresponses}. In contrast, VisConf calibrates distributional confidence within the candidate set, aggregates confidence and visual engagement separately over each rollout before fusion, and uses RWC to retain answer-level consensus.

%%%%%%%%%%%%%%%%%%%%%%%%%%%%%%
\begin{figure}[h]
    \centering

    \begin{subfigure}[t]{0.4\linewidth}
        \centering
        \includegraphics[width=\linewidth]{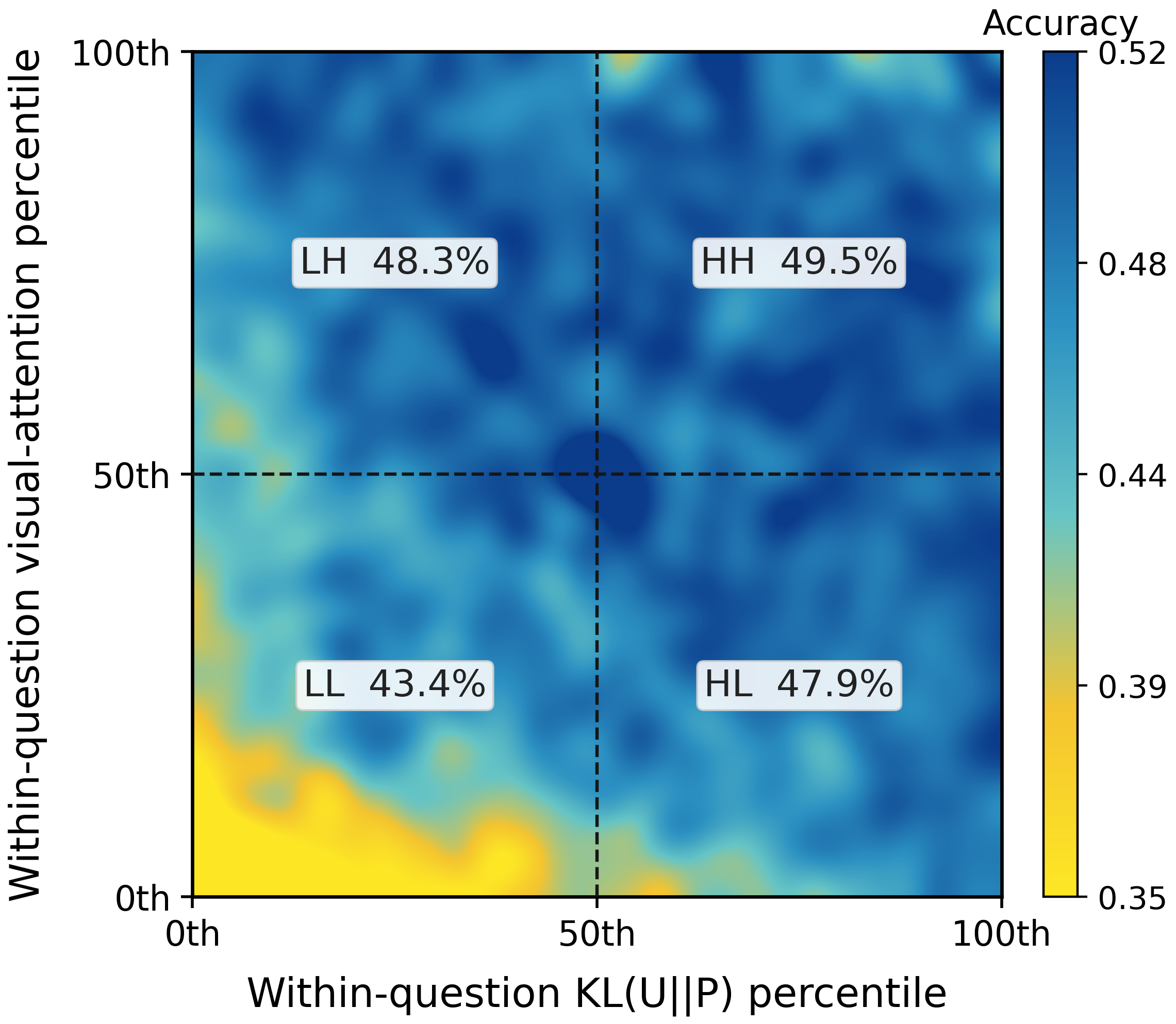}
        \caption{Confidence-vision accuracy landscape}
        \label{fig:acc-landscape}
    \end{subfigure}
    \hfill
    \begin{subfigure}[t]{0.57\linewidth}
        \centering
        \includegraphics[width=\linewidth]{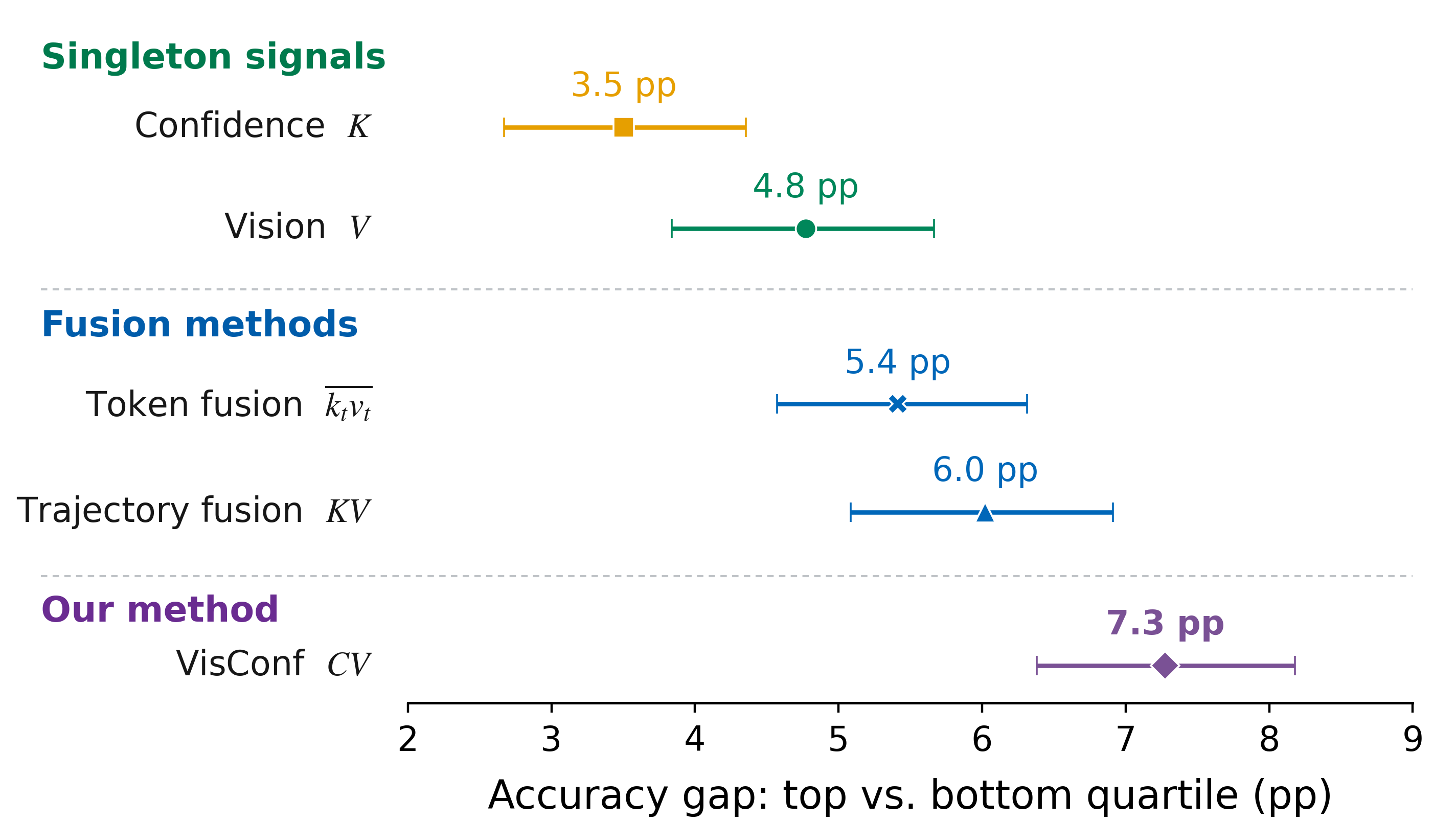}
        \caption{Top-bottom quartile accuracy gap}
        \label{fig:discrim-gain}
    \end{subfigure}

    \caption{Analysis on Qwen2.5-VL-3B using MathVista, MMMU-Pro, and MMStar. (a) Accuracy as a function of confidence and visual engagement. (b) Discriminative strength measured by the accuracy gap between the top and bottom score quartiles. Combining confidence and visual engagement is more informative, but how the signals are combined also matters.
}
    \label{fig:confidence-vision-landscape-discrim-gain}
\end{figure}

%%%%%%%%%%%%%%%%%%%%%%%%%%%%%%%%%%%%%%%%%%%%%%%%%%%%%%%%%%%%
\section{What Signals VLM Rollout Quality?}
BoN succeeds only when the selection rule can distinguish reliable reasoning from plausible errors. During autoregressive generation, the transformer architecture exposes two distinct sources of information at each decoding step: its predictive distribution over the vocabulary, from which we derive \emph{confidence}, and its internal attention to the visual input, which we term \emph{visual engagement}.

\subsection{Standalone Quality Signals}
\paragraph{Confidence via Self-Certainty}
For a generation trajectory \(i\) at step \(t\), let \(p_{i,t}\) denote the next-token distribution obtained by applying softmax directly to the unscaled model logits, and \(U\) the uniform distribution over the vocabulary. The token-level confidence signal is
\[
k_{i,t} = D_{\mathrm{KL}}(U \,\|\, p_{i,t})
\]
where larger values indicate more concentrated predictive distributions \citep{kang2025scalablebestofnselectionlarge}.

\paragraph{Visual Engagement via Visual Attention}
We use visual attention as a token-level measure of visual engagement, following LookBack \citep{cho2026lookbackscorelvlmresponses}. Let \(\alpha_{i,t,j}^{(\ell,h)}\) be the post-softmax attention from the position predicting token \(t\) to context position \(j\) in layer \(\ell\) and head \(h\). We average attention across all layers and heads and sum within each token group \(S\):
\[
\bar{\alpha}_{i,t,j}
=
\frac{1}{\sum_{\ell}H_\ell}
\sum_{\ell}\sum_{h=1}^{H_\ell}\alpha_{i,t,j}^{(\ell,h)},
\qquad
A^S_{i,t}=\sum_{j\in S}\bar{\alpha}_{i,t,j}.
\]
For image (\(\mathcal I\)), prompt-text (\(\mathcal P\)), and all previously generated-token (\(\mathcal G_t\)) positions, excluding padding and non-image special tokens, we define token-level visual engagement as
\[
v_{i,t}
=
\frac{A^{\mathcal I}_{i,t}}
{A^{\mathcal I}_{i,t}+A^{\mathcal P}_{i,t}+A^{\mathcal G_t}_{i,t}}.
\]
Larger \(v_{i,t}\) indicates a greater share of attention allocated to the image relative to textual context.

We then average the token-level signals for a generation \(i\) to obtain raw confidence and visual engagement scores:
\[
K_i=\frac{1}{T_i}\sum_{t=1}^{T_i} k_{i,t},
\qquad
V_i=\frac{1}{T_i}\sum_{t=1}^{T_i} v_{i,t}.
\]
Top-1 BoN selection chooses rollout \(\hat{i}\) among \(N\) rollouts using \(\hat{i}=\arg\max_{i\in[1,N]} Q_i\), where \(Q_i=K_i\) for confidence-based selection and \(Q_i=V_i\) for visual engagement-based selection.

\subsection{Each Signal is Informative But Weak}
We first examine how strongly confidence and visual engagement individually predict correctness. We measure how well trajectory-level scores separate correct from incorrect rollouts using the accuracy gap between the top and bottom score quartiles. A useful quality signal should concentrate more correct rollouts in its highest-scoring quartile. We additionally report macro-averaged prompt-level AUC in the appendix (Table~\ref{tab:rollout-quality-auc}).

Mean confidence, \(K_i\), produces an average top-bottom accuracy gap of \(3.5\) percentage points  (Figure~\ref{fig:discrim-gain}). High confidence may reflect a coherent and internally consistent reasoning path, but VLMs can also be confidently wrong when language priors dominate or an early visual misinterpretation propagates through otherwise fluent reasoning. Mean visual engagement, \(V_i\), yields a slightly larger gap of \(4.8\) points. Sustained image attention may indicate continued use of visual evidence, but can also reflect uncertainty, repeated reinspection, or attention to irrelevant content. Conversely, some correct rollouts may also extract necessary visual information during early reasoning and therefore not require prolonged visual engagement.

\subsection{Confidence and Vision are Complementary at the Trajectory Level}
\paragraph{Joint Signal Structure} The two signals become notably informative when considered jointly. We stratify rollouts by whether their within-prompt raw-confidence and visual engagement scores fall above or below the median. Rollouts high in both signals achieve \(49.5\%\) accuracy, compared with \(48.3\%\) when only visual engagement is high, \(47.9\%\) when only confidence is high, and \(43.4\%\) when both are low (Figure~\ref{fig:acc-landscape}). Holding either signal fixed, higher values of the other are also associated with greater correctness enrichment (Appendix Figure~\ref{fig-app:fix-one-vary-other}). Together, these results indicate that confidence and visual engagement provide complementary rather than redundant evidence of rollout quality.

\paragraph{Trajectory-Level vs.\ Token-Level Fusion}
The joint analysis motivates combining confidence and visual engagement, but does not specify where fusion should occur. Token-level fusion averages the stepwise products, \(Q_i^{\mathrm{tok}}=\frac{1}{T_i}\sum_t v_{i,t}k_{i,t}\), coupling the two signals at the same decoding steps. Trajectory-level fusion instead multiplies their rollout means, \(Q_i^{\mathrm{traj}}=V_iK_i\), allowing confidence and visual engagement to contribute at different stages of generation. Both improve separation over either singleton signal, but trajectory-level fusion produces a larger top--bottom quartile accuracy gap than token-level fusion (\(6.0\) vs.\ \(5.4\) percentage points; Figure~\ref{fig:discrim-gain}).

The two constructions differ only through within-rollout temporal covariance: trajectory-level fusion retains the marginal strength of each signal without requiring their tokenwise peaks to coincide. Consistent with this decomposition, randomizing token correspondence matches the AUC of trajectory-level fusion, whereas aligned token fusion is slightly weaker (Appendix~\ref{app:token-trajectory-fusion}). This suggests that the useful interaction between confidence and visual engagement lies primarily in their trajectory-level magnitudes.
%%%%%%%%%%%%%%%%%%%%%%%%%%%%%%%%%%%%%%%%%%%%%%%%%%%%%%%%%%%%

%%%%%%%%%%%%%%%% METHOD FIGURE %%%%%%%%%%%%%%%%%%%%%%%%%%%%%
\begin{figure}
    \centering
    \includegraphics[width=1\linewidth]{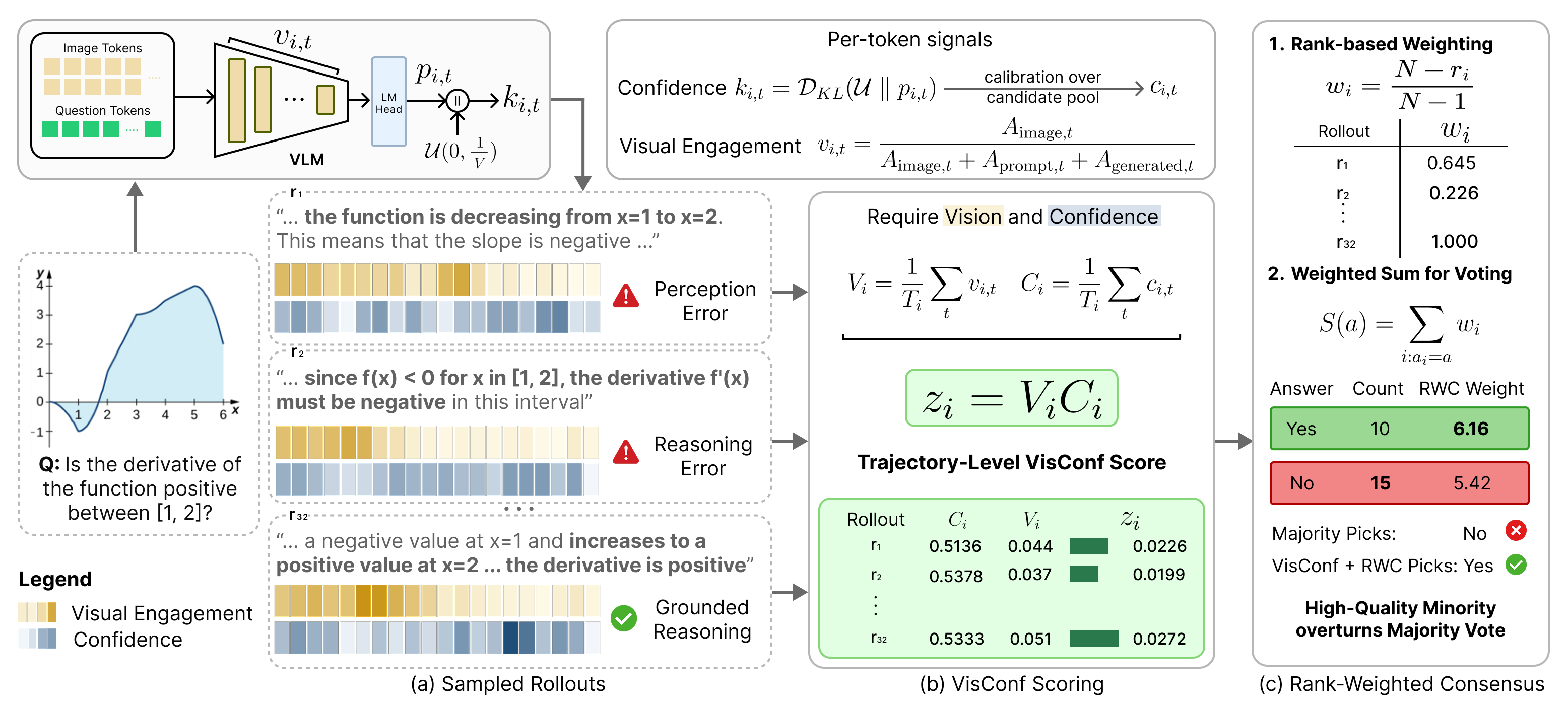}
    \caption{Overview of the VisConf and RWC framework. Given multiple sampled VLM rollouts, VisConf aggregates calibrated confidence and visual engagement independently over each trajectory, and combines them into a rollout-quality score, rewarding trajectories that are both confident and visually engaged. RWC uses VisConf scores to effectively select high-quality answer groups.
}
    \label{fig:main-method-diagram}
\end{figure}
%%%%%%%%%%%%%%%%%%%%%%%%%%%%%%%%%%%%%%%%%%%%%%%%%%%%%%%%%%%%

%%%%%%%%%%%%%%%%%%%%%%%%%%%%%%%%%%%%%%%%%%%%%%%%%%%%%%%%%%%%
\section{VisConf: Trajectory-Level Visual-Confidence Scoring}
Shaped by our analysis in Section 3, we introduce VisConf, which first calibrates confidence relative to the candidate set and then combines rollout-level confidence and visual engagement (Figure~\ref{fig:main-method-diagram}). 

\subsection{Candidate-Relative Confidence Calibration}
To integrate Self-Certainty into a multi-signal score, we calibrate confidence relative to the pooled token-level values across all candidate rollouts \(G\) for a query.
\[
c_{i,t} = \operatorname{clip}_{[0,1]} \left( \frac{k_{i,t}-Q_{.05}(G)}{\max\left(Q_{.95}(G)-Q_{.05}(G),\, \epsilon\right)} \right),
\]
where \(Q_{p}(G)\) is the \(p\)-quantile of pooled token KL values in \(G\), and \(\epsilon > 0\) ensures numerical stability. This scales confidence to \([0,1]\) while bounding the influence of extreme KL values. Using shared percentiles across \(G\) therefore situates each rollout in the context of the candidate set, making subsequent trajectory-level scores relative rather than absolute.

\subsection{Trajectory-Level Visual-Confidence Fusion}
VisConf aggregates visual engagement and calibrated confidence independently over a complete rollout by taking the average across the full generation, resulting in \(V_i\) and \(C_i\), respectively. We then define the VisConf quality score as:
\[
Z_i=V_iC_i.
\]
VisConf favors trajectories that exhibit both high calibrated confidence and high visual engagement, without explicitly requiring that the two signals coincide at the same decoding steps.
%%%%%%%%%%%%%%%%%%%%%%%%%%%%%%%%%%%%%%%%%%%%%%%%%%%%%%%%%%%%

%%%%%%%%%%%%%%%%%%%%%%%%%%%%%%%%%%%%%%%%%%%%%%%%%%%%%%%%%%%%
\section{From Rollout Quality to Answer Consensus}
VisConf estimates the relative quality of each rollout, but selecting only the highest-scoring candidate treats this noisy score as a verifier and ignores corroborating evidence from other trajectories. The value of retaining both sources of evidence is reflected in their joint association with rollout correctness (Appendix Figure~\ref{fig-app:rank-support-heatmap}). Inspired by Borda voting \citep{kang2025scalablebestofnselectionlarge}, we present \emph{Rank-Weighted Consensus} (RWC) to consolidate consensus with quality, as illustrated in Figure~\ref{fig:main-method-diagram}(c).

\subsection{Rank-Based Quality Weighting}
Let \(r_i\) denote the rank of a quality score \(Z_i\) among \(N\) rollouts, with \(r_i=1\) being the highest-ranked candidate. We convert each rank into a normalized quality weight:
\[
w_i=\frac{N-r_i}{N-1},
\]
with exact score ties receiving their average rank. The highest-ranked rollout receives weight \(1\) and the lowest receives \(0\). Lower-ranked rollouts contribute proportionally less. Because these weights depend only on ordering, they are invariant to strictly increasing transformations of the VisConf score. Unlike Self-Certainty’s parameterized Borda weights \(b_p(r_i)=(N-r_i+1)^p\), RWC uses a fixed linear schedule that downweights weak evidence without requiring \(p\) to be tuned \citep{kang2025scalablebestofnselectionlarge}.

\subsection{Quality-Weighted Answer Consensus}
Let \(a_i\) denote the normalized answer extracted from rollout \(i\). We sum the quality weights of all trajectories supporting each answer and select the answer with the highest quality-informed support:
\[
\operatorname{Vote}(a)
=
\sum_{i=1}^{N}\mathbf{1}[a_i=a]\,w_i,
\qquad
\widehat a
=
\arg\max_a\operatorname{Vote}(a).
\]
Thus, repeated answers accumulate support, but higher-quality trajectories exert more influence on the final prediction. A minority answer can therefore prevail when its quality compensates for its lower frequency.
%%%%%%%%%%%%%%%%%%%%%%%%%%%%%%%%%%%%%%%%%%%%%%%%%%%%%%%%%%%%

%%%%%%%%%%%%%%%%%%%%%%%%%%%%%%%%%%%%%%%%%%%%%%%%%%%%%%%%%%%%
\section{Experimental Setup}
We evaluate response-selection methods in VLMs under a shared evaluation protocol.

\textbf{Models and Benchmarks}: We evaluate Qwen2.5-VL-3B \citep{bai2025qwen25vltechnicalreport}, InternVL3-8B \citep{zhu2025internvl3exploringadvancedtraining}, and Gemma4-E2B \citep{gemmateam2026gemma4technicalreport} on three multimodal reasoning benchmarks: MathVista \citep{lu2024mathvistaevaluatingmathematicalreasoning}, MMMU-Pro \citep{yue2025mmmuprorobustmultidisciplinemultimodal}, and MMStar \citep{chen2024rightwayevaluatinglarge}, using the MathVista \texttt{testmini} split and the full evaluation sets of MMMU-Pro and MMStar.

\textbf{Candidate Generation and Evaluation}: For each question, we generate 32 chain-of-thought rollouts \citep{wei2023chainofthoughtpromptingelicitsreasoning, zhang2024multimodalchainofthoughtreasoninglanguage} using a fixed stochastic decoding configuration across models. We use temperature = \(0.7\), top-\(p=0.8\), top-\(k=20\), and a maximum of \(2048\) generated tokens. We report selection accuracy for \(N\in\{8,16,32\}\). For \(N=8\) and \(N=16\), results are averaged over 100 fixed subsets sampled without replacement from the 32-rollout pool using fixed seeds. \(N=32\) uses the complete pool once. Accuracy is averaged within each question before dataset-level aggregation.

\textbf{Baselines}: We compare against \emph{FirstAns}, which is the expected accuracy of one randomly sampled rollout, and \emph{Self-Consistency}, which selects the most frequent answer. We evaluate \emph{Self-Certainty} using top-1 BoN selection as well as Borda-based aggregation with \(p\in \{{0.3,0.7,1.2,2.0}\}\)\footnote{Results for \(p=0.7\) and \(p=2.0\) are reported in Appendix~\ref{app:per-model-dataset-N-results}.} \citep{kang2025scalablebestofnselectionlarge}. For multimodal scoring, we evaluate \emph{LookBack} \citep{cho2026lookbackscorelvlmresponses} using its original top-\(1\) selection and additionally pair its scores with RWC, allowing comparison under the same answer-aggregation rule. Finally, we evaluate VisConf with top-\(1\) selection and RWC, recomputing confidence normalization and ranks within each sampled candidate subset.

%%%%%%%%%%%%%%%%%%%%%%%%%%%%%%%%%%%%%%%%%%%%%%%%%%%%%%%%%%%%

%%%%%%%%%%%%%%% MAIN RESULTS TABLE %%%%%%%%%%%%%%%%%%%%%%%%%
\begin{table*}[t]
\caption{
Averaged accuracy (\%) across MathVista, MMMU-Pro, and MMStar at varying rollout budgets \(N\). Per-dataset results are presented in Appendix~\ref{app:per-model-dataset-N-results}. VisConf + RWC achieves the highest average for every model and sampling budget.}
\label{tab:main-results}
\centering
\small
\setlength{\tabcolsep}{2.5pt}
\renewcommand{\arraystretch}{1.05}

\begin{tabular}{lcccccccccc}
\toprule
\textbf{Method}
& \multicolumn{3}{c}{\textbf{Qwen2.5-VL-3B}}
& \multicolumn{3}{c}{\textbf{InternVL3-8B}}
& \multicolumn{3}{c}{\textbf{Gemma4-E2B}}
& \multirow{2}{*}{\textbf{Avg.}} \\
\cmidrule(lr){2-4}
\cmidrule(lr){5-7}
\cmidrule(lr){8-10}
& \(N=8\) & \(N=16\) & \(N=32\)
& \(N=8\) & \(N=16\) & \(N=32\)
& \(N=8\) & \(N=16\) & \(N=32\)
& \\
\midrule

FirstAns
& 48.51 & 48.51 & 48.51
& 56.55 & 56.55 & 56.55
& 47.34 & 47.34 & 47.34
& 50.80 \\

\midrule

Self-Consistency
& 53.70 & 54.75 & 55.07
& 60.96 & 61.71 & 61.80
& 49.85 & 50.23 & 50.35
& 55.38 \\

Self-Certainty
& 49.78 & 49.95 & 49.94
& 58.83 & 59.03 & 59.07
& 49.43 & 49.62 & 49.81
& 52.83 \\

  + Borda (\(p=0.3\))
& 53.83 & 54.87 & 55.14
& 61.21 & 61.97 & 62.25
& 50.12 & 50.50 & 50.71
& 55.62 \\

  + Borda (\(p=1.2\))
& 53.53 & 54.72 & 55.33
& 61.21 & 62.03 & 62.78
& 50.35 & 50.82 & 50.91
& 55.74 \\

\addlinespace[1pt]

LookBack
& 50.09 & 50.34 & 50.57
& 58.22 & 58.11 & 57.84
& 49.16 & 49.74 & 50.27
& 52.70 \\

  + RWC
& 53.68 & 54.93 & 55.72
& 61.27 & 62.16 & 62.65
& 50.30 & 50.86 & 51.19
& 55.86 \\

\midrule

VisConf
& 51.04 & 51.16 & 51.17
& 59.46 & 59.81 & 60.27
& 50.07 & 50.49 & 50.58
& 53.78 \\

  + RWC
& \textbf{54.09}
& \textbf{55.32}
& \textbf{56.29}
& \textbf{61.69}
& \textbf{62.53}
& \textbf{63.27}
& \textbf{50.89}
& \textbf{51.44}
& \textbf{51.68}
& \textbf{56.36} \\

\bottomrule
\end{tabular}
\end{table*}
%%%%%%%%%%%%%%%%%%%%%%%%%%%%%%%%%%%%%%%%%%%%%%%%%%%%%%%%%%%%

%%%%%%%%%%%%%%%%%%%%%%%%%%%%%%%%%%%%%%%%%%%%%%%%%%%%%%%%%%%%
\section{Results}
\subsection{Overall Performance and Scaling}
Table~\ref{tab:main-results} reports accuracy averaged across the three benchmarks. VisConf + RWC achieves the highest accuracy for every model and sampling budget, with an overall average of \(56.36\%\). This exceeds LookBack + RWC, the strongest baseline overall, by \(0.5\) points; the best Self-Certainty + Borda configuration by \(0.62\) points; and Self-Consistency by \(0.98\) points. Across the per-dataset results (Appendix Table~\ref{tab-app:full-results}), VisConf + RWC outperforms every baseline in \(24\) of \(27\) model-dataset-budget settings and trails by at most \(0.27\) points in the remaining three. Improvement against the strongest baseline remains statistically significant (paired \(95\%\) CI: \([0.32,\,0.66]\)) and positive at every sampling budget (Appendix~\ref{app:statistical-analysis}).

Comparing top-1 selection without aggregation, VisConf achieves an overall average of \(53.78\%\), \(1.08\) and \(0.95\) points higher than LookBack (\(52.70\%\)) and Self-Certainty (\(52.83\%\)), respectively. VisConf’s advantage over LookBack suggests that independently high confidence and visual engagement across decoding stages yields a stronger quality signal than requiring both signals to coincide token-wise. These top-1 results encourage the potential use of VisConf where aggregation schemes are not applicable. In our settings, however, omitting answer-group information is generally restrictive, as top-1 VisConf selection performs below VisConf + RWC across all \(27\) settings. This validates the use of quality-informed aggregation in tasks where outcome-level rollout grouping is available.

\begin{figure}[t]
    \centering
    \begin{subfigure}[t]{0.32\linewidth}
        \centering
        \includegraphics[width=\linewidth]{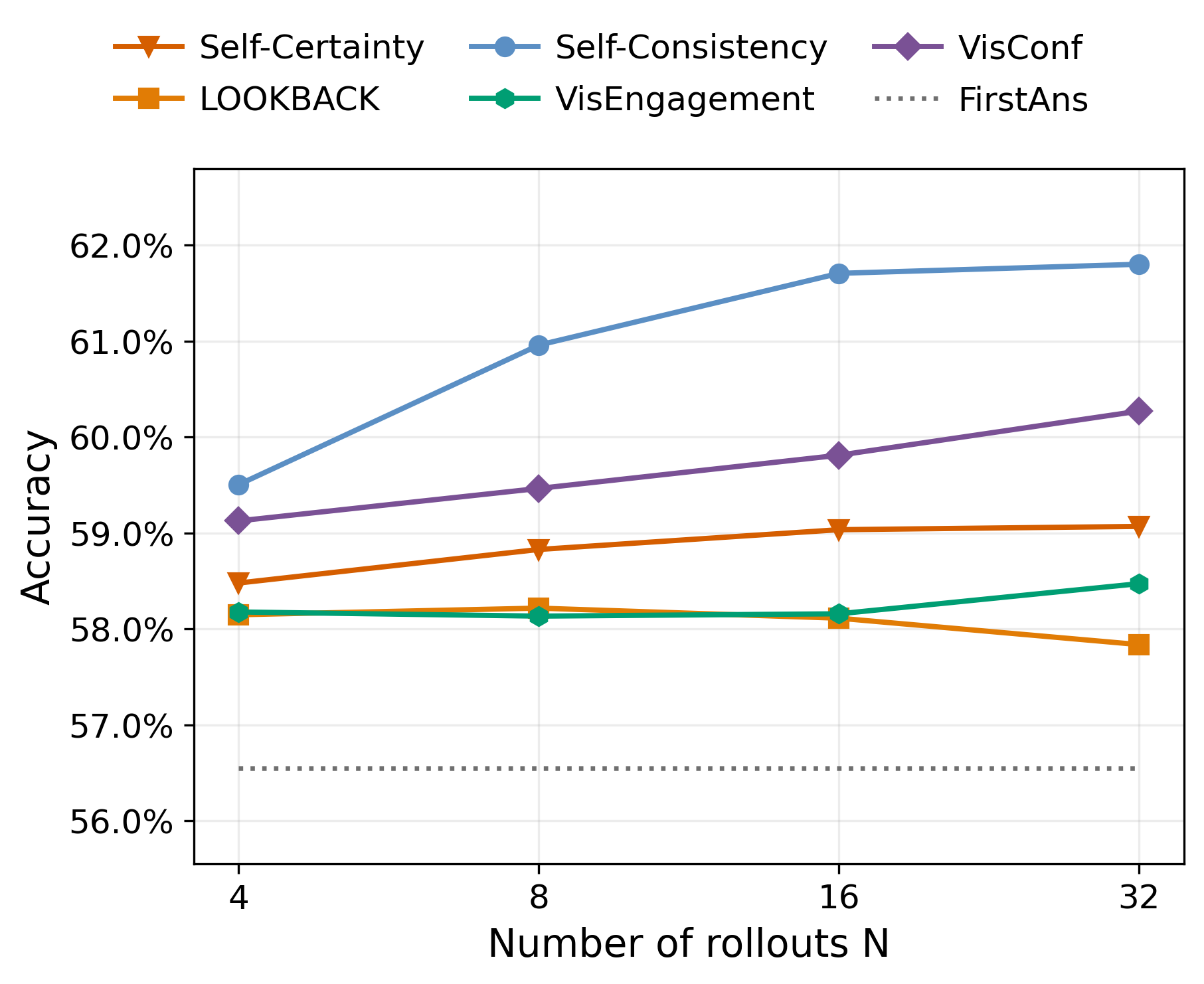}
        \caption{Top-1 Selection: InternVL3-8B}
        \label{fig:internvl-scaling-bon}
    \end{subfigure}
    \hfill
    \begin{subfigure}[t]{0.32\linewidth}
        \centering
        \includegraphics[width=\linewidth]{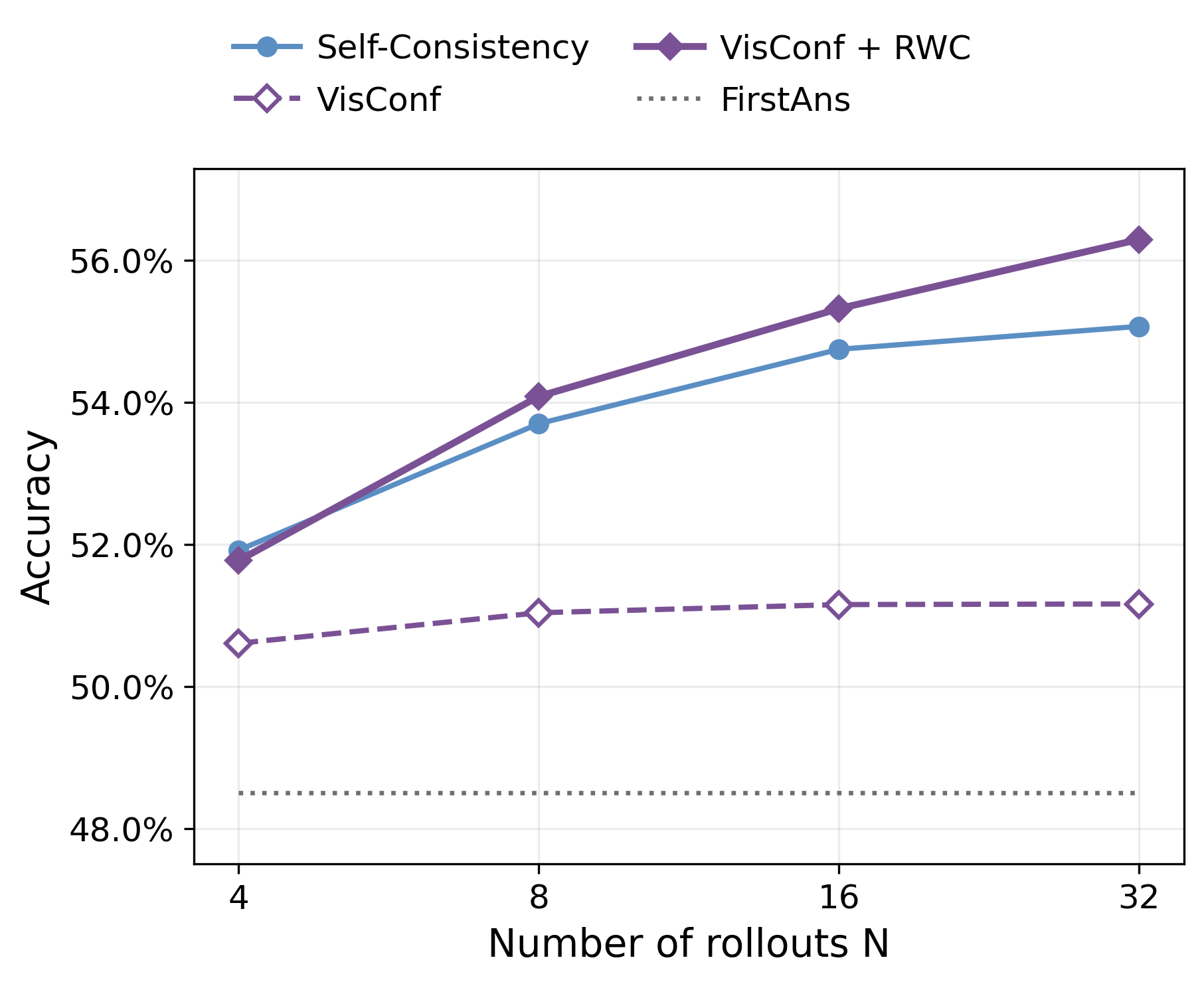}
        \caption{Aggregated: Qwen2.5-VL-3B}
        \label{fig:qwen25-scaling}
    \end{subfigure}
    \hfill
    \begin{subfigure}[t]{0.32\linewidth}
        \centering
        \includegraphics[width=\linewidth]{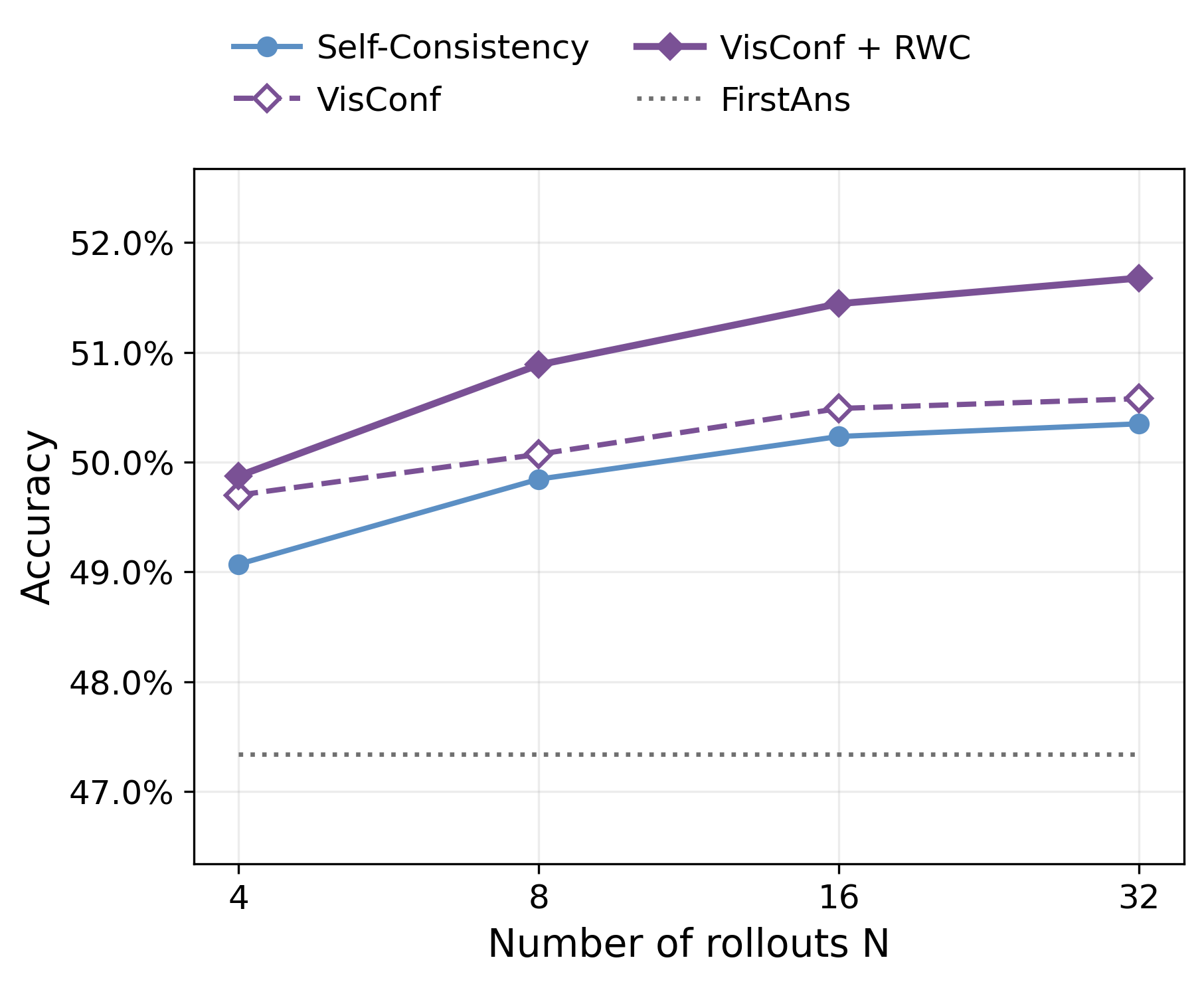}
        \caption{Aggregated: Gemma4-E2B}
        \label{fig:gemma4-scaling}
    \end{subfigure}
    \caption{Effect of scaling \(N\) on accuracy across three benchmarks. (a) VisConf top-1 selection scales strongly but underperforms Self-Consistency. (b-c) VisConf + RWC consistently outperforms top-1 and Self-Consistency, with its advantage persisting or increasing with scaling.}
    \label{fig:avg-scaling-rwc}
\end{figure}

VisConf + RWC provides additional benefits as \(N\) is scaled. From \(N=8\) to \(N=32\), VisConf + RWC improves from \(54.09\%\) to \(56.29\%\) on Qwen2.5-VL-3B, \(61.69\%\) to \(63.27\%\) on InternVL3-8B, and \(50.89\%\) to \(51.68\%\) on Gemma4-E2B (Figure~\ref{fig:avg-scaling-rwc}). Our formulation at \(N=16\) outperforms Self-Consistency at \(N=32\) in all three models, indicating more effective use of the rollout budget.

%%%%%%%%%%%%%%% SCALING FIGURES %%%%%%%%%%%%%%%%%%%%%%%%%
%%%%%%%%%%%%%%%%%%%%%%%%%%%%%%%%%%%%%%%%%%%%%%%%%%%%%%%%%%%%
\noindent
\begin{minipage}[t]{0.52\linewidth} 
\subsection{Ablation Study}
\paragraph{Score Ablation}

Table~\ref{tab:visconf-ablation} shows the results of isolating each VisConf score component under top-1 selection on Qwen2.5-VL-3B. Visual engagement and confidence without fusion are relatively weak selectors, at \(49.58\%\) and \(49.89\%\), while their raw trajectory-level product improves performance to \(50.12\%\). Incorporating candidate-relative calibration with token-level fusion achieves \(50.44\%\), lower than our VisConf trajectory-level formulation, which is \(0.68\) points higher at \(51.12\%\). These results validate individual components of VisConf as meaningfully contributing to the overall performance. Our VisConf score can be generally written as
\[
Z_{i,\alpha}=V_i^{\alpha}C_i^{1-\alpha},
\]
where \(\alpha\) controls the relative influence of visual engagement and confidence. Since \(Z_{i,0.5}=\sqrt{V_iC_i}\), it induces exactly the same ranking as the product \(V_iC_i\) used by VisConf. A post-hoc sensitivity sweep validates this balanced choice without per-dataset tuning (Appendix Figure~\ref{fig-app:visconf-alpha-sens-n32}).

Furthermore, to assess whether VisConf's performance can be attributed to generation length bias, we include a MinLength baseline that selects the shortest rollout. Its substantially lower accuracy (\(49.20\%\)) suggests that VisConf's gains are not driven by a simple preference for shorter generations.

\paragraph{Aggregation Ablation}
Holding the quality score fixed and replacing RWC with raw-score weighted consensus WC (where \(w_i = Z_i\)) reduces accuracy from \(55.23\%\) to \(55.02\%\): ranks preserve the candidate ordering while discarding unreliable score magnitudes.

\end{minipage}
\hfill
\begin{minipage}[t]{0.46\linewidth}
\centering
\small
\captionof{table}{Ablation study on Qwen2.5-VL-3B.
Average accuracy \(\%\) across the three benchmarks for \(N \in \{8,16,32\}\).
}
\label{tab:visconf-ablation}

\vspace{-0.4em}

\setlength{\tabcolsep}{6pt}
\renewcommand{\arraystretch}{1.08}

\begin{tabular}{lc}
\toprule
\textbf{Method} & \textbf{Avg.} \\
\midrule

\rowcolor{gray!12}
\multicolumn{2}{c}{\textbf{Score Ablation with BoN}} \\

Visual Engagement: \(V\)
& 49.58 \\

Self-Certainty: \(K\)
& 49.89 \\

VisConf w/o KL Calibration: \(VK\)
& 50.12 \\

Token Fusion: \(\overline{v_{i,t}c_{i,t}}\)
& 50.44 \\

\textbf{VisConf: \(VC\)}
& \textbf{51.12} \\

MinLength
& 49.20 \\

\midrule

\rowcolor{gray!12}
\multicolumn{2}{c}{\textbf{Aggregation Ablation}} \\

VisConf + WC 
& 55.02 \\

\textbf{VisConf + RWC}
& \textbf{55.23} \\

\bottomrule
\end{tabular}

\vspace{1em}
\includegraphics[width=\linewidth]{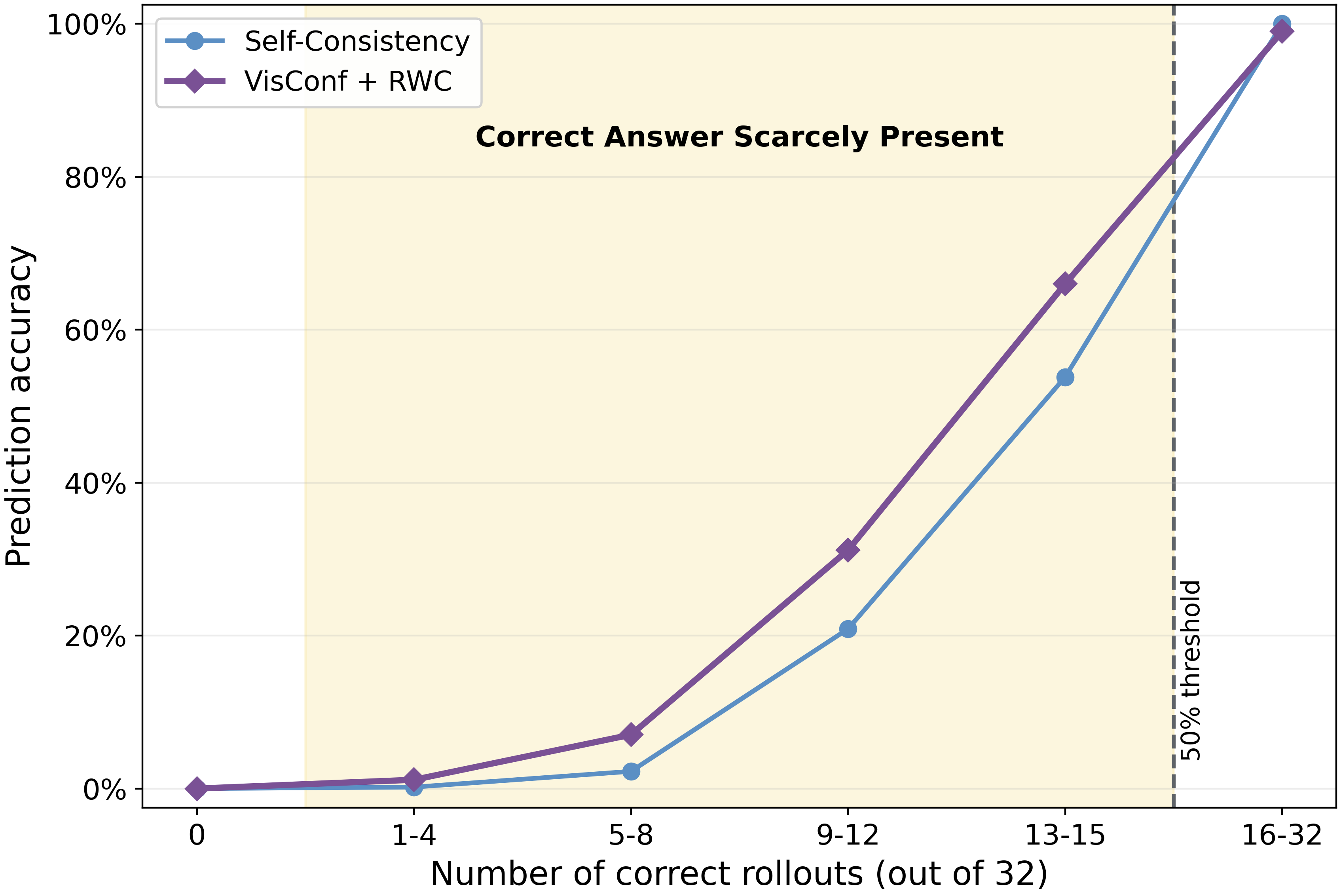}
    \captionof{figure}{Performance comparison against Self-Consistency on InternVL3-8B in cases where the candidate pool contains few correct rollouts. VisConf + RWC effectively prioritizes correct evidence without majority support.}
    \label{fig:accuracy_by_gold_evidence_sc_vs_ours}
\end{minipage}

\subsection{Performance When Correct Evidence Is Sparse}
Figure~\ref{fig:accuracy_by_gold_evidence_sc_vs_ours} examines how performance varies with empirical generation difficulty, measured by the prevalence of correct-answer trajectories in the \(N=32\) candidate pool,
\[
g_q=\frac{n_{\mathrm{correct}}}{N}.
\]
VisConf + RWC increasingly outperforms Self-Consistency as correct evidence approaches, but remains below, the absolute majority threshold. Its advantage rises from \(0.96\) percentage points with \(1\)-\(4\) correct rollouts to \(4.79\) points with \(5\)-\(8\), \(10.29\) points with \(9\)-\(12\), and \(12.18\) points with \(13\)-\(15\). All \(110\) questions rescued relative to Self-Consistency occur within \(0<g_q<0.5\), compared with \(29\) regressions, yielding \(81\) net corrections in this subset (Appendix Table~\ref{app-tab:rescues-regressions-by-gold-evidence}). Thus, VisConf + RWC provides its largest gains when substantial correct evidence exists but remains outnumbered under unweighted consensus, using quality information to prioritize stronger evidence over frequency alone.

%%%%%%%%%%%%%%%%%%%%%%%%%%%%%%%%%%%%%%%%%%%%%%%%%%%%%%%%%%%%

\section{Discussion}
\textbf{Best-of-N is both a discovery and an aggregation problem.} Increasing the rollout budget results in two competing outcomes: it increases the chance of discovering a correct reasoning path, but is likely to introduce additional incorrect candidates which may also be high-scoring. Our direct-selection results demonstrate that simply extending the candidate set does not guarantee better BoN selection, even when the oracle accuracy of the pool continues to increase (Appendix \ref{app:per-model-dataset-N-results}). The stronger scaling of VisConf + RWC suggests that the value of additional test-time compute depends not only on generating better candidates, but also on preserving useful evidence distributed across the candidate set. We observe that this preservation can positively affect outcomes in cases where correct reasoning is less commonly encountered.  

\textbf{Relative quality is more reliable than absolute quality.}
Confidence and visual engagement are noisy proxies, and neither of them provides a reliable estimate that a rollout is correct. VisConf and RWC therefore use these signals comparatively while carefully avoiding spurious conclusions. Candidate-relative calibration contextualizes confidence within the current pool, multiplicative fusion promotes the occurrence of both signals, and RWC retains only ordinal rankings rather than interpreting raw magnitudes. This separation of roles is important. VisConf provides a general measure of which trajectories appear stronger, while RWC gauges how this imperfect ordering aligns with answer-level agreement. An extreme raw score's domination is therefore subdued (Appendix~\ref{app:rwc-bounded-influence}) and any single trajectory's outcome is not treated as conclusive. Our work suggests that effective selection is possible without requiring a perfect verifier. Weak model-internal signals can become useful provided that the influence of noise is systematically reduced.

\textbf{Limitations and Future Work.} While we incorporate attention as a reference for the model engaging with input image(s), attention alone does not establish correct visual understanding or faithfully explain the model’s reasoning \citep{jain2019attentionexplanation, wiegreffe-pinter-2019-attention}. Obtaining visual-attention signals also requires eager attention extraction. In a controlled replay study on Qwen2.5-VL-3B, the median runtime overhead relative to FlashAttention-2 was negligible at \(N=8\) and \(N=16\) but increased to \(18.4\%\) at \(N=32\) (Appendix \ref{app:attn-ext-overhead}). Future work could replace attention with a cheaper proxy or one that more causally explains whether visual evidence is reliably used. Although VisConf's superior top-1 performance over other scoring baselines suggests that it may remain useful when answer-level grouping is unavailable, we do not evaluate this setting directly. More broadly, VisConf provides a training-free estimate of rollout quality that need not be confined to selection, and future work could study its use as a reward signal for policy optimization or in other paradigms where judging trajectory quality is useful.

\section{Conclusion}

We introduced VisConf, a simple, training-free response quality score using model-internal quality signals. Together with RWC, our framework combines rollout quality with answer-level agreement to improve BoN selection while requiring no tunable hyperparameters or external reward models. Our results demonstrate that imperfect model-internal signals can support effective test-time scaling when used comparatively and accumulated across agreeing trajectories rather than treated as standalone verifiers.

%%%%%%%%%%%%%%%%%%%%%%%%%%%%%%%%%%%%%%%%%%%%%%%%%%%%%%%%%%%%

%%%%%%%%%%%%%%%%%%%%%%%%%%%%%%%%%%%%%%%%%%%%%%%%%%%%%%%%%%%%

\bibliographystyle{plainnat}
\bibliography{references}

%%%%%%%%%%%%%%%%%%%%%%%%%%%%%%%%%%%%%%%%%%%%%%%%%%%%%%%%%%%%
%%%%%%%%%%%%%%%%%%%%%%%%%%%%%%%%%%%%%%%%%%%%%%%%%%%%%%%%%%%%
\appendix
%%%%%%%%%%%%%%%%%%%%%%%%%%%%%%%%%%%%%%%%%%%%%%%%%%%%%%%%%%%%
%%%%%%%%%%%%%%%%%%%%%%%%%%%%%%%%%%%%%%%%%%%%%%%%%%%%%%%%%%%%

%%%%%%%%%%%%%%%%%%%%%%%%%%%%%%%%%%%%%%%%%%%%%%%%%%%%%%%%%%%%
\section{Experimental and Implementation Details}
\paragraph{Prompt templates}
Each prompt contains the image(s), question, and answer choices when applicable. Qwen2.5-VL and InternVL3 receive the following chain-of-thought prompt: ``Think step by step inside \texttt{\textless think\textgreater\ \textless/think\textgreater} tags, then give your final answer inside \texttt{\textbackslash boxed\{\}}.'' Gemma4 is run with thinking mode disabled and receives the same instruction without explicit thinking tags: ``Think step by step, then give your final answer inside \texttt{\textbackslash boxed\{\}}.'' Inputs are rendered using each model's native chat template.

\paragraph{Dataset information}

We evaluate the 1,000-example MathVista \texttt{testmini} split, the complete 1,730-example MMMU-Pro standard test set, and the complete 1,500-example MMStar validation set for all three models. 

MathVista evaluates mathematical reasoning in visual contexts and contains both multiple-choice and free-form questions. It contains arithmetic, algebraic, geometric, logical, numerical, statistical, and scientific reasoning problems, with visual inputs including diagrams, charts, tables, function plots, scientific figures, natural images, and puzzle-style scenes \citep{lu2024mathvistaevaluatingmathematicalreasoning}.

MMMU-Pro targets college-level reasoning across a broad range of disciplines, including science, engineering, medicine, business, humanities, and the social sciences. Questions require combining domain knowledge with information from various visual inputs such as diagrams, plots, tables, maps, and scientific figures. The benchmark is constructed from MMMU by filtering questions that can be answered reliably without the image, making visual evidence more necessary for solving the task \citep{yue2025mmmuprorobustmultidisciplinemultimodal}.

MMStar consists of visually dependent multiple-choice questions organized across six capability categories: coarse perception, fine-grained perception, instance reasoning, logical reasoning, science and technology, and mathematics. These are further divided into 18 fine-grained dimensions covering tasks such as object localization and counting, attribute recognition, cross-instance comparison, diagram reasoning, numerical calculation, and statistical analysis. Its examples are curated to reduce language-only shortcuts and ensure that visual information is necessary for answering the question \citep{chen2024rightwayevaluatinglarge}.

\paragraph{Answer extraction and normalization}
We extract the last well-formed \(\backslash\texttt{boxed}\{\cdot\}\) expression in each response. Multiple-choice answers are canonicalized to uppercase option letters. Boxed option text is matched to its corresponding choice. When no valid boxed output is present, conservative final-sentence and final-line patterns are used. Free-form MathVista answers are compared using normalized numerical or string matching, including fraction handling and decimal comparison. Unextractable responses are assigned a shared \texttt{<NO\_ANSWER>} label.

\paragraph{Tie handling} Direct Best-of-N selection resolves score ties using the lowest original rollout index. For RWC, equal VisConf scores receive their average rank; if all scores are equal, every rollout receives weight \(0.5\). If answer-level votes are tied, we select the answer whose earliest supporting rollout has the lowest original index.

\paragraph{Compute Resources} Experiments with Qwen2.5-VL-3B and Gemma4-E2B used one NVIDIA GeForce RTX 4090 GPU with 24 GB of memory; InternVL3-8B experiments used one NVIDIA A100 GPU with 80 GB of memory.

\paragraph{LookBack Implementation}
\begin{table*}[h!]
\centering
\caption{LookBack Best-of-N and Rank-Weighted Consensus (RWC)
performance under two values of \(\alpha\), with \(\lambda=1.25\).}
\label{app-tab:lookback-alpha-comparison}

\vspace{-0.3em}
\setlength{\tabcolsep}{4.5pt}
\renewcommand{\arraystretch}{1.08}

\resizebox{\textwidth}{!}{%
\begin{tabular}{cclcccccccccc}
\toprule
\(\boldsymbol{\alpha}\)
& \(\boldsymbol{\lambda}\)
& \textbf{Aggregation}
& \multicolumn{3}{c}{\textbf{Qwen2.5-VL-3B}}
& \multicolumn{3}{c}{\textbf{InternVL3-8B}}
& \multicolumn{3}{c}{\textbf{Gemma4-E2B}}
& \textbf{Avg.} \\
\cmidrule(lr){4-6}
\cmidrule(lr){7-9}
\cmidrule(lr){10-12}
& & 
& \(N=8\) & \(N=16\) & \(N=32\)
& \(N=8\) & \(N=16\) & \(N=32\)
& \(N=8\) & \(N=16\) & \(N=32\)
& \\ 
\midrule

0.25 & 1.25 & BoN
& 49.88 & 50.17 & 50.45
& 56.31 & 55.00 & 52.99
& 48.92 & 49.38 & 49.95
& 51.45 \\

0.25 & 1.25 & RWC
& 53.40 & 54.61 & 55.51
& 60.78 & 61.68 & 62.07
& 50.16 & 50.70 & 50.84
& 55.53 \\

\midrule

0.50 & 1.25 & BoN
& 50.09 & 50.34 & 50.57
& 58.22 & 58.11 & 57.84
& 49.16 & 49.74 & 50.27
& 52.70 \\

0.50 & 1.25 & RWC
& 53.68 & 54.93 & 55.72
& 61.27 & 62.16 & 62.65
& 50.30 & 50.86 & 51.19
& 55.86 \\

\bottomrule
\end{tabular}%
}
\end{table*}
For LookBack, let \(\ell_{i,t}\) be the sampled-token log-probability and \(A_{i,t}\) the all-layer, all-head image-attention mass. Its response score is
\[
L_i=
\frac{\sum_t A_{i,t}^{\lambda}
\left(\ell_{i,t}+\alpha\log A_{i,t}\right)}
{\sum_t A_{i,t}^{\lambda}}.
\]
The work suggests a choice of \(\alpha=0.25\) and \(\lambda=1.25\) as a global setting. However, we observe that \(\alpha=0.5\) and \(\lambda=1.25\), which were the model-specific settings for Qwen2.5-VL-7B, achieve greater performance (Table \ref{app-tab:lookback-alpha-comparison}). Thus, we fix \(\alpha=0.5\) and \(\lambda=1.25\) across all models, datasets, and sampling budgets in our baseline results. Standard LookBack selects the highest-scoring rollout. LookBack + RWC applies our rank-weighted answer aggregation to the same \(L_i\) scores.

%%%%%%%%%%%%%%%%%%%%%%%%%%%%%%%%%%%%%%%%%%%%%%%%%%%%%%%%%%%%

%%%%%%%%%%%%%%%%%%%%%%%%%%%%%%%%%%%%%%%%%%%%%%%%%%%%%%%%%%%%

%%%%%%%%%%%%%%%%%%%%%%%%%%%%%%%%%%%%%%%%%%%%%%%%%%%%%%%%%%%%
\section{Full Results and Statistical Analysis}
\subsection{Per-Model, Dataset, and \(N\) Results}
\label{app:per-model-dataset-N-results}
\begin{table*}[h!]
\caption{Full accuracy results (\%) on MathVista, MMMU-Pro, and MMStar for
\(N\in\{8,16,32\}\). Avg. is the unweighted mean across the nine dataset-budget combinations within each model.}
\label{tab-app:full-results}
\centering
\small
\setlength{\tabcolsep}{2.5pt}
\renewcommand{\arraystretch}{1.05}

\begin{tabular}{lcccccccccc}
\toprule
\textbf{Method}
& \multicolumn{3}{c}{\textbf{MathVista}}
& \multicolumn{3}{c}{\textbf{MMMU-Pro}}
& \multicolumn{3}{c}{\textbf{MMStar}}
& \multirow{2}{*}{\textbf{Avg.}} \\
\cmidrule(lr){2-4}
\cmidrule(lr){5-7}
\cmidrule(lr){8-10}
& \(N=8\) & \(N=16\) & \(N=32\)
& \(N=8\) & \(N=16\) & \(N=32\)
& \(N=8\) & \(N=16\) & \(N=32\)
& \\
\midrule

% ================================================================
% Qwen2.5-VL-3B
% ================================================================
\rowcolor{gray!12}
\multicolumn{11}{c}{\textbf{Qwen2.5-VL-3B}} \\
\addlinespace[2pt]

Oracle
& 82.76 & 87.01 & 89.40
& 80.31 & 88.10 & 92.95
& 85.50 & 91.57 & 95.67
& 88.14 \\

FirstAns
& 54.01 & 54.01 & 54.01
& 41.20 & 41.20 & 41.20
& 50.31 & 50.31 & 50.31
& 48.51 \\

\midrule

Self-Consistency
& 61.04 & 62.48 & 63.20
& 45.30 & 46.38 & 46.88
& 54.77 & 55.38 & 55.13
& 54.51 \\

Self-Certainty
& 55.65 & 55.70 & 55.40
& 42.97 & 43.25 & 43.76
& 50.72 & 50.89 & 50.67
& 49.89 \\

+ Borda (\(p=0.3\))
& \textbf{61.35} & 62.78 & 63.60
& 45.52 & 46.55 & 46.76
& 54.63 & 55.27 & 55.07
& 54.61 \\

+ Borda (\(p=0.7\))
& 61.25 & 62.79 & 63.90
& 45.60 & 46.64 & 47.28
& 54.48 & 55.08 & 54.93
& 54.66 \\

+ Borda (\(p=1.2\))
& 60.88 & 62.61 & 63.90
& 45.55 & 46.58 & 47.23
& 54.15 & 54.96 & 54.87
& 54.53 \\

+ Borda (\(p=2.0\))
& 60.07 & 62.26 & 63.90
& 45.28 & 46.42 & 47.23
& 53.68 & 54.72 & 55.07
& 54.29 \\

\addlinespace[1pt]

LookBack
& 55.56 & 55.64 & 55.50
& 42.98 & 43.22 & 43.41
& 51.74 & 52.17 & 52.80
& 50.33 \\

+ RWC
& 60.66 & 62.44 & 63.70
& 45.75 & 46.90 & 47.34
& 54.64 & 55.44 & \textbf{56.13}
& 54.78 \\

\midrule

VisConf
& 56.63 & 56.82 & 57.10
& 43.57 & 43.44 & 43.06
& 52.93 & 53.21 & 53.33
& 51.12 \\

+ RWC
& 61.08 & \textbf{62.97} & \textbf{64.50}
& \textbf{46.12} & \textbf{47.24} & \textbf{48.38}
& \textbf{55.06} & \textbf{55.75} & 56.00
& \textbf{55.23} \\

% ================================================================
% InternVL3-8B
% ================================================================
\midrule
\rowcolor{gray!12}
\multicolumn{11}{c}{\textbf{InternVL3-8B}} \\
\addlinespace[2pt]

Oracle
& 80.46 & 83.92 & 86.50
& 80.69 & 86.67 & 90.64
& 84.77 & 88.81 & 91.80
& 86.03 \\

FirstAns
& 57.57 & 57.57 & 57.57
& 49.43 & 49.43 & 49.43
& 62.65 & 62.65 & 62.65
& 56.55 \\

\midrule

Self-Consistency
& 63.62 & 64.66 & 64.60
& 53.28 & 54.02 & 54.34
& 65.97 & 66.44 & 66.47
& 61.49 \\

Self-Certainty
& 60.75 & 60.79 & 60.50
& 51.68 & 52.15 & 52.37
& 64.05 & 64.16 & 64.33
& 58.98 \\

+ Borda (\(p=0.3\))
& 64.05 & 65.12 & 65.70
& 53.55 & 54.24 & 54.45
& 66.03 & 66.55 & 66.60
& 61.81 \\

+ Borda (\(p=0.7\))
& 64.15 & 65.25 & 66.10
& 53.66 & 54.38 & 54.62
& 66.06 & 66.57 & 66.73
& 61.95 \\

+ Borda (\(p=1.2\))
& 63.99 & 65.19 & 66.40
& 53.62 & 54.43 & 55.14
& 66.03 & 66.49 & 66.80
& 62.01 \\

+ Borda (\(p=2.0\))
& 63.58 & 64.91 & 66.20
& 53.41 & 54.33 & 55.20
& 65.83 & 66.45 & 66.60
& 61.84 \\

\addlinespace[1pt]

LookBack
& 59.46 & 59.34 & 59.80
& 50.85 & 50.82 & 50.17
& 64.34 & 64.17 & 63.53
& 58.05 \\

+ RWC
& 63.71 & 65.00 & 65.90
& 53.72 & 54.60 & 54.97
& 66.38 & 66.88 & 67.07
& 62.02 \\

\midrule

VisConf
& 61.04 & 61.26 & 61.40
& 52.25 & 52.90 & 53.41
& 65.10 & 65.27 & 66.00
& 59.85 \\

+ RWC
& \textbf{64.29} & \textbf{65.49} & \textbf{67.00}
& \textbf{54.13} & \textbf{54.96} & \textbf{55.55}
& \textbf{66.65} & \textbf{67.14} & \textbf{67.27}
& \textbf{62.50} \\

% ================================================================
% Gemma4-E2B
% ================================================================
\midrule
\rowcolor{gray!12}
\multicolumn{11}{c}{\textbf{Gemma4-E2B}} \\
\addlinespace[2pt]

Oracle
& 74.64 & 80.08 & 83.50
& 72.81 & 79.87 & 84.97
& 79.38 & 84.96 & 88.73
& 80.99 \\

FirstAns
& 47.63 & 47.63 & 47.63
& 41.90 & 41.90 & 41.90
& 52.49 & 52.49 & 52.49
& 47.34 \\

\midrule

Self-Consistency
& 50.14 & 50.56 & 50.90
& 44.29 & 44.59 & 44.68
& 55.10 & 55.55 & 55.47
& 50.14 \\

Self-Certainty
& 49.21 & 49.43 & 50.60
& 43.77 & 43.87 & 43.76
& 55.30 & 55.55 & 55.07
& 49.62 \\

+ Borda (\(p=0.3\))
& 50.34 & 50.76 & 50.90
& 44.49 & 44.83 & 45.38
& 55.54 & 55.90 & 55.87
& 50.44 \\

+ Borda (\(p=0.7\))
& 50.44 & 50.89 & 50.80
& 44.62 & 45.03 & 45.49
& 55.75 & 56.18 & 56.13
& 50.59 \\

+ Borda (\(p=1.2\))
& 50.49 & 50.90 & 50.60
& 44.69 & 45.16 & 45.32
& 55.87 & 56.38 & 56.80
& 50.69 \\

+ Borda (\(p=2.0\))
& 50.44 & 50.87 & 50.80
& 44.71 & 45.23 & 45.78
& 55.94 & 56.54 & \textbf{56.87}
& 50.80 \\

\addlinespace[1pt]

LookBack
& 48.98 & 49.40 & 49.80
& 44.00 & 44.97 & 46.01
& 54.48 & 54.85 & 55.00
& 49.72 \\

+ RWC
& 50.35 & 50.76 & 51.10
& 45.00 & 45.66 & 46.07
& 55.54 & 56.15 & 56.40
& 50.78 \\

\midrule

VisConf
& 49.95 & 50.54 & 50.80
& 44.87 & 45.25 & 45.20
& 55.39 & 55.68 & 55.73
& 50.38 \\

+ RWC
& \textbf{50.93} & \textbf{51.48} & \textbf{51.70}
& \textbf{45.64} & \textbf{46.22} & \textbf{46.53}
& \textbf{56.09} & \textbf{56.63} & 56.80
& \textbf{51.34} \\

\bottomrule
\end{tabular}
\end{table*}

\subsection{Paired Statistical Analysis}
\begin{table*}[h!]
\caption{
Paired accuracy improvements of VisConf + RWC over each comparator in
percentage points with \(95\%\) question-cluster bootstrap
confidence intervals. Overall averages across sampling budgets, models, and benchmarks.
}
\label{tab:paired-significance}
\centering
\footnotesize
\setlength{\tabcolsep}{5pt}
\renewcommand{\arraystretch}{1.08}

\begin{tabular}{lcccc}
\toprule
\textbf{Comparator}
& \(\boldsymbol{N=8}\)
& \(\boldsymbol{N=16}\)
& \(\boldsymbol{N=32}\)
& \textbf{Overall} \\
\midrule

LookBack + RWC
& \(+0.47\;[0.35,\,0.59]\)
& \(+0.45\;[0.31,\,0.61]\)
& \(+0.56\;[0.24,\,0.87]\)
& \(+0.49\;[0.32,\,0.66]\) \\

Self-Consistency
& \(+0.72\;[0.56,\,0.86]\)
& \(+0.87\;[0.68,\,1.07]\)
& \(+1.34\;[0.97,\,1.72]\)
& \(+0.98\;[0.76,\,1.18]\) \\

Self-Cert. + Borda (\(p=1.2\))
& \(+0.52\;[0.40,\,0.66]\)
& \(+0.58\;[0.40,\,0.74]\)
& \(+0.74\;[0.39,\,1.09]\)
& \(+0.61\;[0.43,\,0.80]\) \\

\midrule

VisConf (top-\(1\))
& \(+2.03\;[1.79,\,2.28]\)
& \(+2.61\;[2.21,\,3.02]\)
& \(+3.08\;[2.42,\,3.77]\)
& \(+2.57\;[2.15,\,3.00]\) \\

\bottomrule
\end{tabular}
\end{table*}
\label{app:statistical-analysis}

We estimate uncertainty using a paired question-cluster bootstrap (Table~\ref{tab:paired-significance}). For \(N\in\{8,16\}\), accuracy is first averaged over the 100 shared candidate subsets for each question; \(N=32\) uses the complete rollout pool. We resample the 4,230 questions with replacement within each benchmark, keeping evaluations of the same question paired across models. Method differences are computed before aggregation and macro-averaged across the nine model-benchmark combinations. The overall estimate additionally averages across \(N\in\{8,16,32\}\). We report percentile \(95\%\) confidence intervals from 2,000 bootstrap replicates.

%%%%%%%%%%%%%%%%%%%%%%%%%%%%%%%%%%%%%%%%%%%%%%%%%%%%%%%%%%%%

\begin{minipage}[t]{0.53\linewidth} 
\vspace{0pt}
\section{Computational Overhead of Attention Extraction}
\label{app:attn-ext-overhead}
We measure the overhead of replacing FlashAttention-2 with eager attention and explicit attention extraction using a controlled replay experiment on Qwen2.5-VL-3B. We replay the exact stored output-token sequences for nested candidate pools of \(N\in\{8,16,32\}\) on 120 MMMU-Pro prompts, with three repetitions per condition, while holding prompts, images, generated tokens, batching, and KV caching fixed. Figure \ref{fig-app:runtime_ratio} reports the paired wall-clock runtime ratio of eager attention extraction to FlashAttention-2, computed per prompt after taking the median over repetitions. Median runtime is comparable at \(N=8\) and \(N=16\), with ratios of \(0.99\times\) and \(1.01\times\), respectively, while the overhead becomes more pronounced at \(N=32\), where eager attention extraction requires \(1.18\times\) the runtime. Because the benchmark excludes scoring and answer aggregation, these measurements represent the inference-time cost of obtaining the attention signal itself.

\end{minipage}
\hfill
\begin{minipage}[t]{0.45\linewidth}
\vspace{1.2\baselineskip}
\centering
\includegraphics[width=\linewidth]{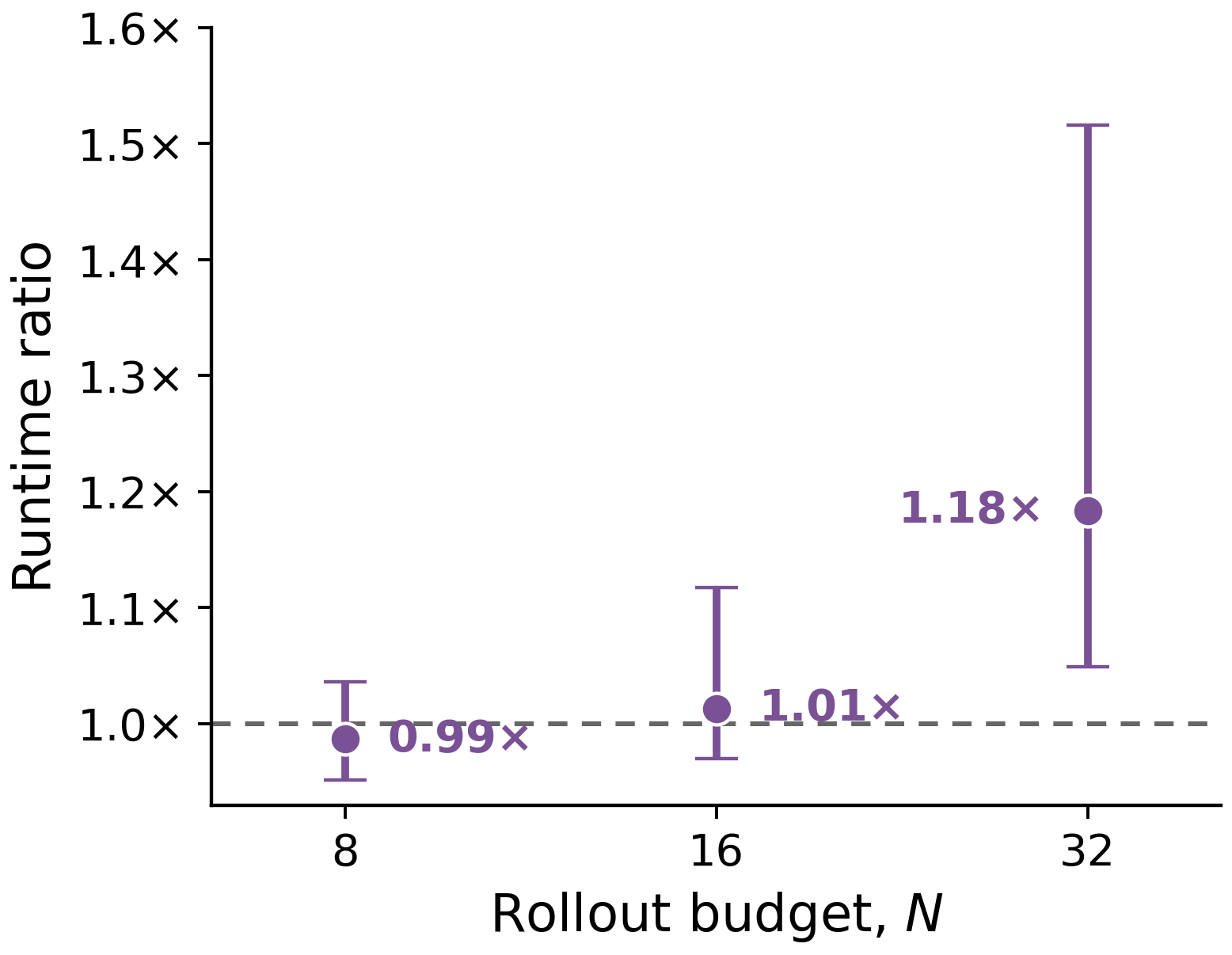}
    \captionof{figure}{Paired wall-clock runtime ratio of eager attention extraction to FlashAttention-2 on Qwen2.5-VL-3B across 120 MMMU-Pro prompts. Points show median ratios and error bars show the interquartile range.}
    \label{fig-app:runtime_ratio}
\end{minipage}

%%%%%%%%%%%%%%%%%%%%%%%%%%%%%%%%%%%%%%%%%%%%%%%%%%%%%%%%%%%%
\section{Analysis of the VisConf Score}
\subsection{Joint Enrichment of Confidence and Visual Engagement}
\begin{figure}[h!]
    \centering
    \begin{subfigure}[t]{0.4\linewidth}
        \centering
        \includegraphics[width=\linewidth]{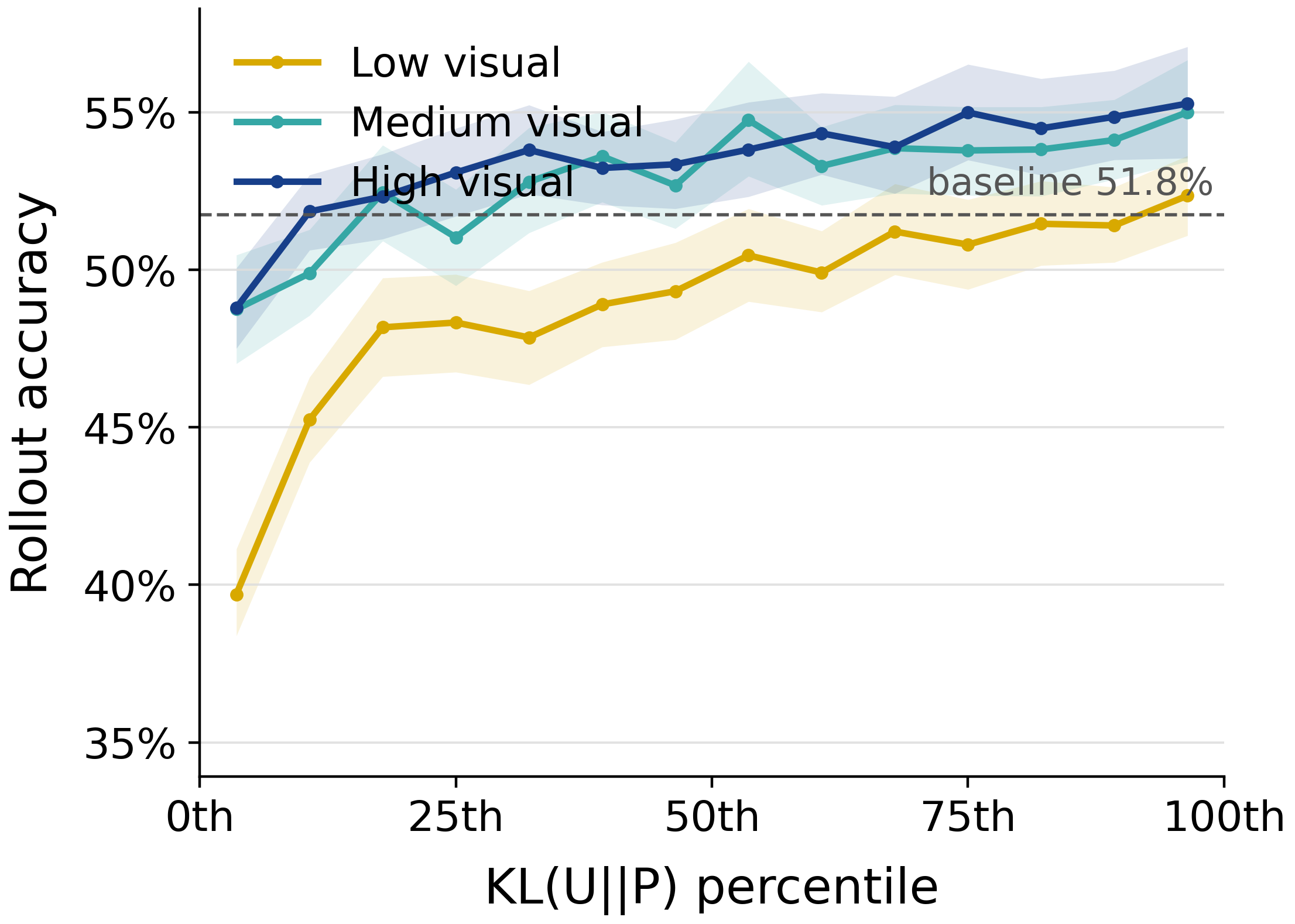}
        \caption{Accuracy across confidence percentiles, stratified by visual engagement tercile.}
        \label{fig-app:fix-kl-vary-attn}
    \end{subfigure}
    \begin{subfigure}[t]{0.4\linewidth}
        \centering
        \includegraphics[width=\linewidth]{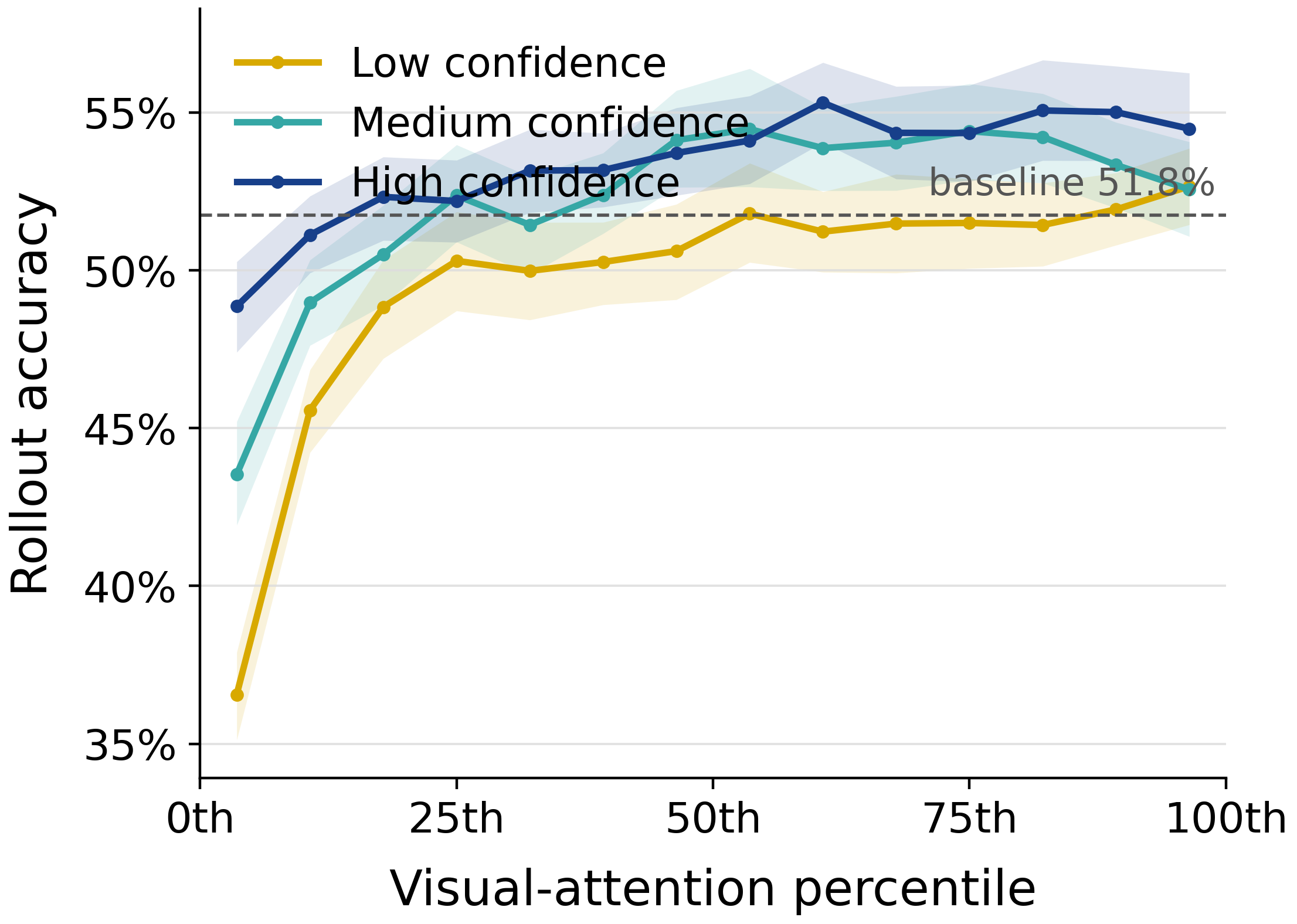}
        \caption{Accuracy across visual engagement percentiles, stratified by confidence tercile.}
        \label{fig-app:fix-attn-vary-kl}
    \end{subfigure}

    \caption{Joint enrichment of rollout correctness by raw confidence and visual engagement. The dashed line denotes overall rollout accuracy, and shaded regions show 95\% confidence intervals.}
    \label{fig-app:fix-one-vary-other}
\end{figure}

Figure~\ref{fig-app:fix-one-vary-other} examines each signal while stratifying by the other. Rollout accuracy generally increases with both confidence and visual engagement, while weak values of either signal limit accuracy. This complementary enrichment motivates combining the two signals in VisConf.

\subsection{Token- and Trajectory-Level Fusion}
\label{app:token-trajectory-fusion}

We analyze the raw confidence signal \(k_{i,t}\), before applying the candidate-relative calibration used by VisConf, to isolate the effect of where confidence and visual engagement are fused. For rollout \(i\) with \(T_i\) generated tokens, define
\begin{equation}
V_i
=
\frac{1}{T_i}\sum_{t=1}^{T_i}v_{i,t}, 
\qquad
K_i
=
\frac{1}{T_i}\sum_{t=1}^{T_i}k_{i,t}.
\label{eq:app-raw-trajectory-means}
\end{equation}
Token-level and trajectory-level fusion are respectively
\begin{equation}
Q_i^{\mathrm{tok}}
=
\frac{1}{T_i}\sum_{t=1}^{T_i}v_{i,t}k_{i,t},
\qquad
Q_i^{\mathrm{traj}}
=
V_iK_i.
\label{eq:app-token-trajectory-definitions}
\end{equation}

\paragraph{Covariance decomposition}
By the definition of within-rollout covariance,
\begin{equation}
Q_i^{\mathrm{tok}}
=
V_iK_i
+
\operatorname{Cov}_{t}(v_{i,t},k_{i,t}).
\label{eq:app-covariance-decomposition}
\end{equation}
Token-level fusion therefore augments the product of the two trajectory means
with a term determined by their temporal co-occurrence. Trajectory-level fusion
retains the marginal strength of each signal without requiring them to be large
at the same decoding positions.

\paragraph{Permutation identity}
Let \(\pi_i\) be a uniformly random permutation of the \(T_i\)
visual engagement positions. Then
\begin{equation}
\mathbb{E}_{\pi_i}
\left[
\frac{1}{T_i}
\sum_{t=1}^{T_i}
v_{i,\pi_i(t)}k_{i,t}
\right]
=
V_iK_i.
\label{eq:app-permutation-identity}
\end{equation}
For every position \(t\), the permuted value \(v_{i,\pi_i(t)}\) is uniformly
distributed over the visual engagement sequence and therefore has expectation
\(V_i\). Consequently,
\begin{equation}
\frac{1}{T_i}
\sum_t
\mathbb{E}_{\pi_i}[v_{i,\pi_i(t)}]\,k_{i,t}\
=
\frac{1}{T_i}\sum_t V_ik_{i,t}
=
K_iV_i.
\end{equation}
Thus, trajectory-level fusion is the expected token product after temporal
correspondence between confidence and visual engagement has been removed. The
same identities hold after replacing raw confidence \(k_{i,t}\) with calibrated
confidence \(c_{i,t}\); VisConf uses the resulting trajectory-level product
\(V_iC_i\).

\paragraph{Within-question discrimination}
We evaluate these formulations on the \(3{,}558\) mixed-correctness questions
from Qwen2.5-VL-3B across MathVista, MMMU-Pro, and MMStar at \(N=32\). A
mixed-correctness question contains at least one correct and one incorrect
rollout.

For question \(q\), let \(\mathcal C_q\) and \(\mathcal I_q\) denote its sets
of correct and incorrect rollouts. For rollout score \(s\), we compute
\begin{equation}
\operatorname{AUC}_q(s)
=
\frac{1}{|\mathcal C_q|\,|\mathcal I_q|}
\sum_{i\in\mathcal C_q}
\sum_{j\in\mathcal I_q}
\mathbf{1}[s_i>s_j].
\label{eq:app-question-auc}
\end{equation}
We macro-average this quantity across questions so that questions with more
correct--incorrect rollout pairs do not receive greater weight. Confidence
intervals are obtained by bootstrap resampling questions. For permuted token
fusion, we average over \(100\) independently sampled within-rollout
permutations.

\begin{center}
\captionof{table}{
Within-question discrimination of rollout-quality signals on
Qwen2.5-VL-3B, pooled across MathVista, MMMU-Pro, and MMStar.
Fusion-location controls use raw confidence \(k\), whereas VisConf uses
candidate-calibrated confidence \(c\).
}
\label{tab:rollout-quality-auc}
\small
\setlength{\tabcolsep}{5pt}
\renewcommand{\arraystretch}{1.08}
\begin{tabular}{llcc}
\toprule
\textbf{Signal}
& \textbf{Score}
& \textbf{AUC}
& \textbf{95\% CI} \\
\midrule
Confidence
& \(K_i\)
& 0.5279
& [0.5214, 0.5344] \\

Visual engagement
& \(V_i\)
& 0.5307
& [0.5243, 0.5374] \\

\midrule
Aligned token fusion
& \(\frac{1}{T_i}\sum_t k_{i,t}v_{i,t}\)
& 0.5373
& [0.5304, 0.5440] \\

Permuted token fusion
& \(\frac{1}{T_i}\sum_t k_{i,t}v_{i,\pi_i(t)}\)
& 0.5399
& [0.5333, 0.5468] \\

Trajectory fusion
& \(K_iV_i\)
& 0.5399
& [0.5331, 0.5462] \\

\midrule
\textbf{VisConf}
& \(\mathbf{C_iV_i}\)
& \textbf{0.5503}
& \textbf{[0.5434, 0.5569]} \\
\bottomrule
\end{tabular}
\end{center}

Randomizing token correspondence produces essentially the same discrimination
as trajectory-level fusion, whereas preserving the original token alignment is
slightly weaker. Tokenwise correspondence therefore provides no additional
discrimination in our analysis.

The covariance term in Eq.~\ref{eq:app-covariance-decomposition} is negatively
oriented with respect to correctness. Its within-question AUC is \(0.4634\),
with a \(95\%\) confidence interval of \([0.4570,0.4697]\). The question-macro
mean covariance is \(-0.02659\) for correct rollouts and \(-0.02507\) for
incorrect rollouts, giving a correct-minus-incorrect difference of
\(-0.00152\), with a \(95\%\) confidence interval of
\([-0.00193,-0.00110]\).

These observations are consistent with the lack of improvement from adding the
observed covariance term. They do not imply that correct reasoning requires
confidence and visual processing to occur at separate stages. Rather, they show
that greater tokenwise co-occurrence may not be a useful quality signal.

\begin{figure}[t]
    \centering

    \begin{minipage}[t]{0.52\linewidth}
        \centering
        \includegraphics[width=\linewidth]{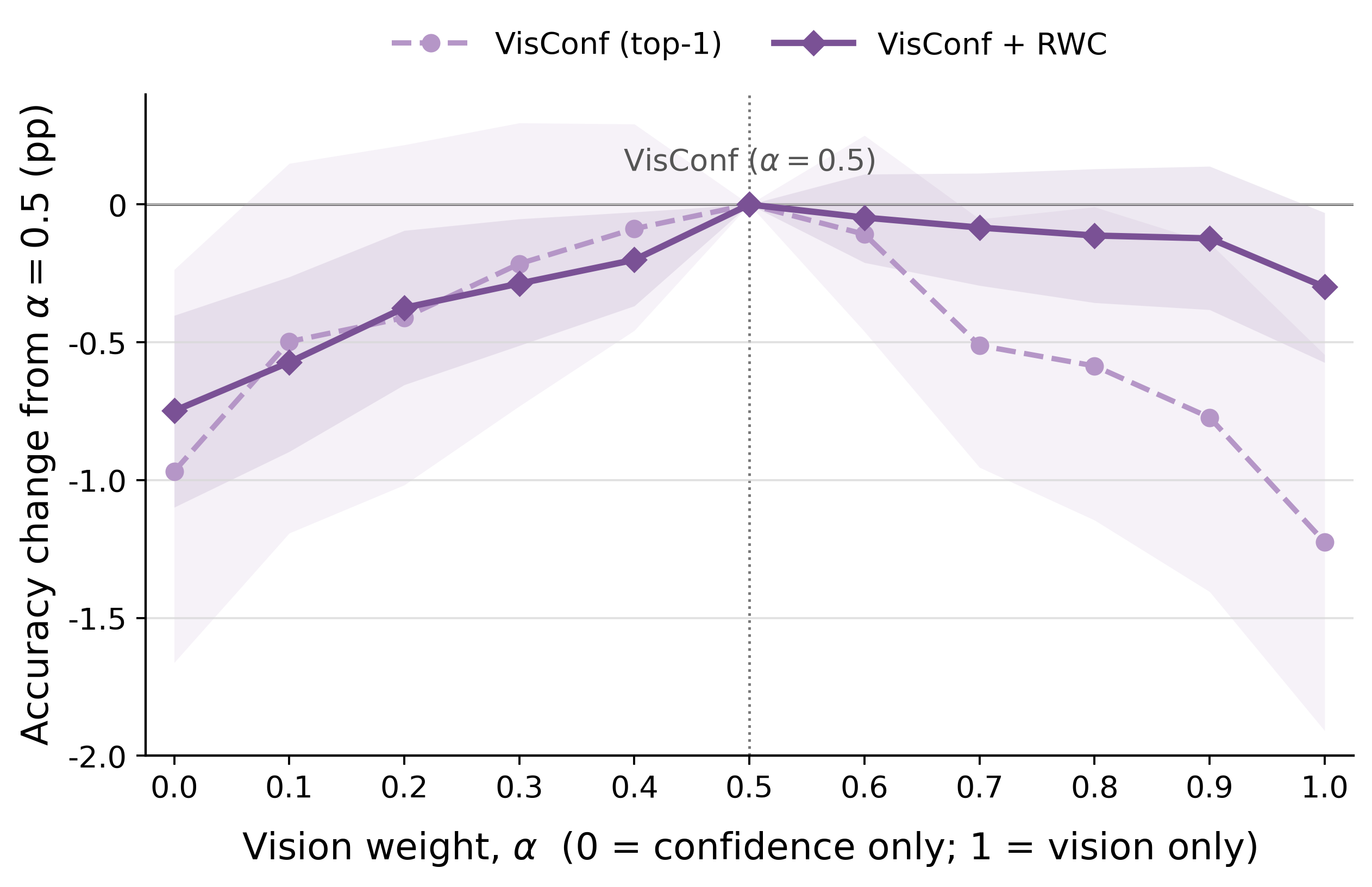}
        \captionof{figure}{
            Post-hoc sensitivity of VisConf to the confidence--vision
            weighting parameter \(\alpha\). Performance is strongest near
            the balanced setting \(\alpha=0.5\).
        }
        \label{fig-app:visconf-alpha-sens-n32}
    \end{minipage}
    \hfill
    \begin{minipage}[t]{0.45\linewidth}
        \centering
        \includegraphics[width=\linewidth]{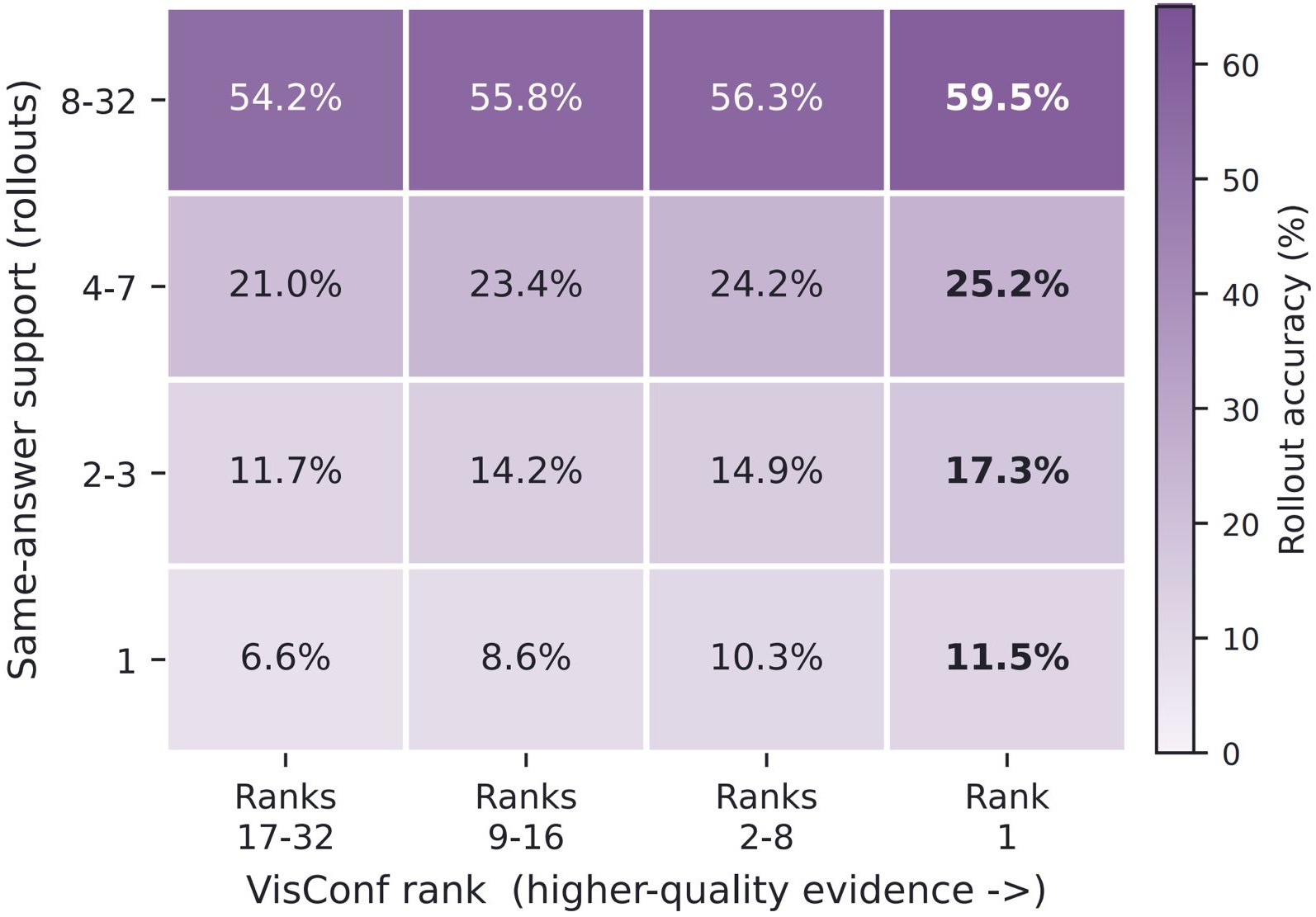}
        \captionof{figure}{
            Answer correctness by same-answer support and VisConf rank. Accuracy increases along both dimensions.
        }
        \label{fig-app:rank-support-heatmap}
    \end{minipage}
\end{figure}
\subsection{Sensitivity to Confidence-Vision Weighting}

To test the sensitivity of VisConf to the relative contribution of confidence and visual engagement, we generalize its multiplicative score as
\[
Z_{i,\alpha}=V_i^{\alpha}C_i^{1-\alpha},
\qquad \alpha\in[0,1].
\]
This weighted geometric form preserves the requirement that both signals contribute while allowing their relative importance to vary. The endpoints \(\alpha=0\) and \(\alpha=1\) recover confidence-only and vision-only scoring, respectively. At \(\alpha=0.5\), \(Z_{i,0.5}=\sqrt{V_iC_i}\), which is rank-equivalent to the VisConf product \(V_iC_i\). The post-hoc sweep in Figure~\ref{fig-app:visconf-alpha-sens-n32} shows that performance is strongest under broadly balanced weighting and varies smoothly around the symmetric choice, indicating that VisConf is not narrowly dependent on \(\alpha=0.5\).

%%%%%%%%%%%%%%%%%%%%%%%%%%%%%%%%%%%%%%%%%%%%%%%%%%%%%%%%%%%%

%%%%%%%%%%%%%%%%%%%%%%%%%%%%%%%%%%%%%%%%%%%%%%%%%%%%%%%%%%%%
\section{Analysis of Rank-Weighted Consensus}
\subsection{Answer Support and Rollout Quality}

RWC combines answer frequency with rollout quality. Figure ~\ref{fig-app:rank-support-heatmap} shows that, at fixed same-answer support, higher VisConf-ranked rollouts are more accurate. Likewise, at fixed rank, greater answer support is associated with higher accuracy. This supports combining both signals in RWC.

\subsection{Rescues and Regressions by Rollout Correctness Frequency}
\label{app:rescue-regression-sec}
\begin{table}[h!]
\centering
\caption{Rescues and regressions of VisConf + RWC relative to
Self-Consistency on InternVL3-8B at \(N=32\), pooled across
MathVista, MMMU-Pro, and MMStar.}
\label{app-tab:rescues-regressions-by-gold-evidence}

\setlength{\tabcolsep}{6pt}
\renewcommand{\arraystretch}{1.08}

\begin{tabular}{lrrrr}
\toprule
\textbf{Correct rollouts}
& \textbf{Questions}
& \textbf{Rescues}
& \textbf{Regressions}
& \textbf{Net} \\
\midrule
\(1\)--\(4\)   & 521   & 6   & 1  & \(+5\)  \\
\(5\)--\(8\)   & 355   & 18  & 1  & \(+17\) \\
\(9\)--\(12\)  & 340   & 52  & 17 & \(+35\) \\
\(13\)--\(15\) & 197   & 34  & 10 & \(+24\) \\
\midrule
\textbf{\(1\)--\(15\)}
& \textbf{1,413}
& \textbf{110}
& \textbf{29}
& \textbf{\(+81\)} \\
\midrule
\(16\)--\(32\) & 2,397 & 0 & 24 & \(-24\) \\
\bottomrule
\end{tabular}
\end{table}

To localize the paired changes underlying the gains in Figure ~\ref{fig:accuracy_by_gold_evidence_sc_vs_ours}, we partition questions by the number of correct rollouts in the \(N=32\) pool. A \emph{rescue} occurs when Self-Consistency is wrong but VisConf + RWC is correct; a \emph{regression} is the reverse. Table~\ref{app-tab:rescues-regressions-by-gold-evidence} reports these transitions on the same InternVL3-8B evaluation used in Figure~\ref{fig:accuracy_by_gold_evidence_sc_vs_ours}.

\subsection{Robustness Properties of Rank-Weighted Consensus}
\label{app:rwc-bounded-influence}

RWC depends only on the ordering of VisConf scores. Consequently, any strictly
increasing transformation of the scores leaves the rollout weights and selected
answer unchanged. This makes aggregation insensitive to the absolute scale of
VisConf.

For \(w_i=(N-r_i)/(N-1)\), the rank weights also satisfy
\[
\sum_{i=1}^{N} w_i=\frac{N}{2},
\qquad
\frac{w_i}{\sum_j w_j}\leq\frac{2}{N}.
\]
Thus, no individual rollout contributes more than \(2/N\) of the normalized
RWC mass. An isolated extreme score therefore cannot dominate the decision
through its magnitude; an answer must instead accumulate support across
high-ranked trajectories.

\subsection{Borda and RWC Aggregation Rules}
\begin{table*}[h!]
\caption{Accuracy (\%) on Qwen2.5-VL-3B, obtained by applying RWC or parameterized Borda
aggregation to the same VisConf rollout scores. Avg is the unweighted
mean across MathVista, MMMU-Pro, and MMStar for
\(N\in\{8,16,32\}\).}
\label{tab:visconf-aggregation}
\centering
\small
\setlength{\tabcolsep}{2.5pt}
\renewcommand{\arraystretch}{1.05}

\begin{tabular}{lcccccccccc}
\toprule
\textbf{Aggregation}
& \multicolumn{3}{c}{\textbf{MathVista}}
& \multicolumn{3}{c}{\textbf{MMMU-Pro}}
& \multicolumn{3}{c}{\textbf{MMStar}}
& \multirow{2}{*}{\textbf{Avg.}} \\
\cmidrule(lr){2-4}
\cmidrule(lr){5-7}
\cmidrule(lr){8-10}
& \(N=8\) & \(N=16\) & \(N=32\)
& \(N=8\) & \(N=16\) & \(N=32\)
& \(N=8\) & \(N=16\) & \(N=32\)
& \\
\midrule

% ================================================================
% Qwen2.5-VL-3B
% ================================================================

RWC
& 61.08 & 62.97 & \textbf{64.50}
& \textbf{46.12} & \textbf{47.24} & \textbf{48.38}
& 55.06 & \textbf{55.75} & 56.00
& \textbf{55.23} \\

Borda (\(p=0.3\))
& \textbf{61.46} & 63.02 & 64.10
& 45.79 & 46.80 & 47.28
& 55.00 & 55.59 & 55.73
& 54.97 \\

Borda (\(p=0.7\))
& \textbf{61.46} & \textbf{63.11} & 64.40
& 46.04 & 47.09 & 48.15
& \textbf{55.10} & 55.69 & 55.87
& 55.21 \\

Borda (\(p=1.2\))
& 61.14 & 62.91 & 64.10
& 46.10 & 47.20 & \textbf{48.38}
& 55.07 & 55.74 & \textbf{56.07}
& 55.19 \\

Borda (\(p=2.0\))
& 60.41 & 62.43 & 63.60
& 45.92 & 47.12 & 48.32
& 54.85 & 55.73 & 56.00
& 54.93 \\

\bottomrule
\end{tabular}
\end{table*}
Parameterized Borda aggregation assigns
\[
b_p(r_i)=(N-r_i+1)^p.
\]
For \(p=1\), this is an affine transformation of the RWC rollout weight:
\[
b_1(r_i)=(N-1)w_i+1.
\]
Consequently, its answer-group score satisfies
\[
S_{p=1}(a)=(N-1)S_{\mathrm{RWC}}(a)+n_a,
\]
Therefore, linear Borda adds an answer-frequency term to the rank-weighted quality evidence. This distinction is relevant to the failure mode studied in Section 7.3 when correct trajectories are high-quality but remain a minority: the additive \(n_a\) term in linear Borda explicitly favors the more frequent answer, partially restoring the majority bias that RWC is intended to overcome. More generally, \(p<1\) further compresses rank differences and approaches unweighted consensus as \(p\to0\), whereas \(p>1\) increasingly concentrates influence on the highest-ranked trajectories. RWC instead uses a fixed linear rank schedule with no additive frequency bonus.

Across the range of evaluated \(p\) values, Borda aggregation is broadly comparable to RWC, reinforcing that rank-weighted answer aggregation is generally effective for combining rollout quality with consensus. RWC offers a simpler alternative by fixing the weighting schedule and removing the need to tune the exponent \(p\), while retaining comparable performance across models, datasets, and sampling budgets.

%%%%%%%%%%%%%%%%%%%%%%%%%%%%%%%%%%%%%%%%%%%%%%%%%%%%%%%%%%%%

%%%%%%%%%%%%%%%%%%%%%%%%%%%%%%%%%%%%%%%%%%%%%%%%%%%%%%%%%%%%

\section{Qualitative Examples}
\label{app:qualitative-examples}

We present five examples at $N=32$ to illustrate distinct qualitative outcomes: two corrections by VisConf + RWC, one regression, one case in which all methods succeed, and one case in which all methods fail. The examples span two models and all three benchmarks. For each example, \emph{support} denotes either the number of votes for the selected answer under Self-Consistency or the sum of normalized within-question rank weights under RWC. LookBack top-1 selects a single rollout and therefore has no answer-group support value.

\newcommand{\qualcorrect}{\textcolor{green!50!black}{correct}}
\newcommand{\qualincorrect}{\textcolor{red!70!black}{incorrect}}
\newcommand{\qualcase}[7]{%
\begin{center}
\setlength{\fboxsep}{8pt}%
\fcolorbox{black!20}{white}{%
\begin{minipage}{0.96\linewidth}
\raggedright
\textbf{#1}

\vspace{6pt}
\begin{minipage}[c]{0.38\linewidth}
    \centering
    \includegraphics[width=\linewidth,height=0.22\textheight,keepaspectratio]{#2}
\end{minipage}\hfill
\begin{minipage}[c]{0.58\linewidth}
    \small
    \textbf{Question.} #3

    \vspace{4pt}
    \textbf{Gold answer:} #4

    \vspace{4pt}
    \textbf{Candidate pool.} #5
\end{minipage}

\vspace{6pt}
\centering
\small
\renewcommand{\arraystretch}{1.08}
\setlength{\tabcolsep}{8pt}
\begin{tabular}{@{}lccc@{}}
    \toprule
    Method & Prediction & Outcome & Support \\
    \midrule
    #6
    \bottomrule
\end{tabular}

\vspace{6pt}
\raggedright
\small
\textbf{Interpretation.} #7
\end{minipage}}
\end{center}
\medskip
}

\subsection{Corrections by VisConf + RWC}

\qualcase
{\textbf{(a)} Visual proportion estimation -- Qwen2.5-VL-3B, MMStar sample 359}
{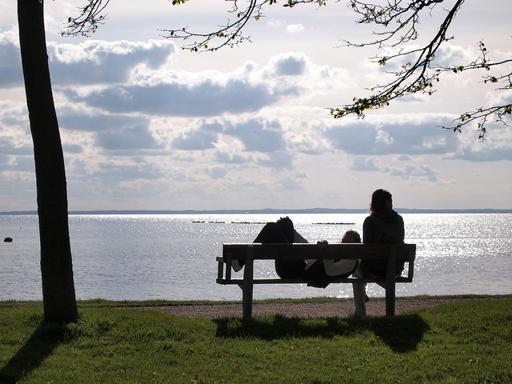}
{Roughly how much of the image is occupied by the two people on the bench?
The relevant options are A: more than $50\%$ and B: less than $30\%$.}
{B (less than $30\%$)}
{A receives 16 rollouts, B receives 15, and D receives one.}
{Self-Consistency & A & \qualincorrect & 16 votes \\
 LookBack top-1 & A & \qualincorrect & -- \\
 LookBack + RWC & A & \qualincorrect & 8.48 \\
 VisConf + RWC & \textbf{B} & \qualcorrect & \textbf{9.06} \\}
{The unweighted vote is wrong by one rollout. Correct trajectories explicitly
compare the people with the much larger water and sky regions, whereas many A
trajectories treat the people as the image's semantic focus and infer that they
must occupy most of its area. VisConf + RWC gives the minority B group enough
collective weight to reverse this narrow incorrect plurality.}

\qualcase
{\textbf{(b)} Tacheometric calculation -- InternVL3-8B, MMMU-Pro sample
\texttt{test\_Architecture\_and\_Engineering\_492}}
{}
{Using the staff readings in the figure, determine the horizontal distance
from A to B and the reduced level of B.}
{D: $AB=94.14\,\mathrm{m}$ and $\mathrm{R.L.}=988.81\,\mathrm{m}$}
{C receives 14 rollouts, D receives 13, A and no-answer receive two each, and
B receives one.}
{Self-Consistency & C & \qualincorrect & 14 votes \\
 LookBack top-1 & C & \qualincorrect & -- \\
 LookBack + RWC & C & \qualincorrect & 7.03 \\
 VisConf + RWC & \textbf{D} & \qualcorrect & \textbf{7.61} \\}
{The candidate pool is fragmented and the correct answer trails the leading
incorrect answer by one rollout. The incorrect C trajectories typically reach
a nearby horizontal-distance value but propagate an error through the
trigonometric or reduced-level calculation. VisConf + RWC shifts the decision
to the D group, whose trajectories more consistently retain the required
quantities through the complete calculation.}

\subsection{A Regression}

\qualcase
{\textbf{(c)} Relational counting -- InternVL3-8B, MathVista sample 984}
{}
{Is the number of gray cars less than the number of small metallic minivans?
The options are A: yes and B: no.}
{B (no)}
{B receives 17 rollouts and A receives 15.}
{Self-Consistency & \textbf{B} & \qualcorrect & 17 votes \\
 LookBack top-1 & \textbf{B} & \qualcorrect & -- \\
 LookBack + RWC & \textbf{B} & \qualcorrect & \textbf{8.58} \\
 VisConf + RWC & A & \qualincorrect & 9.58 \\}
{Here, the calibrated score is misleading. The candidate pool has a narrow but
correct majority for B, and both LookBack variants preserve that decision.
VisConf instead ranks enough of the incorrect A trajectories highly to reverse
the result. This example illustrates that confidence and visual engagement
remain imperfect quality signals: strong engagement with an image does not
guarantee that the depicted objects were counted or categorized correctly.}

\subsection{Agreement and Shared Failure}

\qualcase
{\textbf{(d)} Mood recognition -- Qwen2.5-VL-3B, MMStar sample 13}
{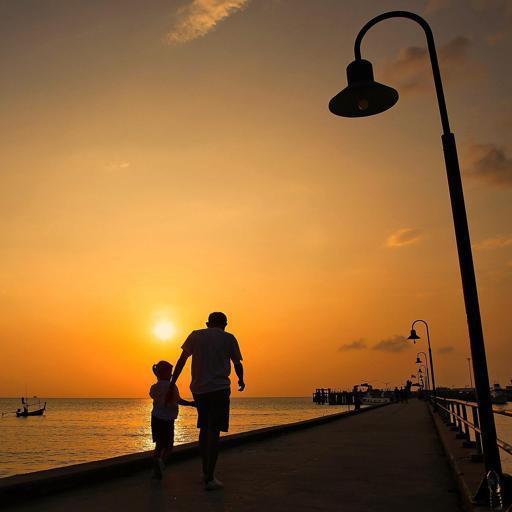}
{Which mood does the image convey? The relevant options are B: cozy and
C: happy.}
{B (cozy)}
{B receives 16 rollouts, C receives 15, and one rollout produces no answer.}
{Self-Consistency & \textbf{B} & \qualcorrect & 16 votes \\
 LookBack top-1 & \textbf{B} & \qualcorrect & -- \\
 LookBack + RWC & \textbf{B} & \qualcorrect & \textbf{9.61} \\
 VisConf + RWC & \textbf{B} & \qualcorrect & \textbf{8.23} \\}
{This is a closely contested but stable agreement case. Both leading groups
recognize the warm sunset and the companionship between the adult and child;
they differ mainly in whether those cues imply ``cozy'' or ``happy.'' Despite
the one-vote margin, both rollout-quality scores and both aggregation rules
retain the correct B decision.}

\qualcase
{\textbf{(e)} Dominant-color recognition -- Qwen2.5-VL-3B, MMStar sample 276}
{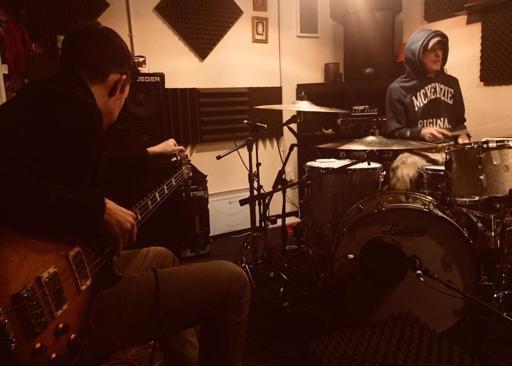}
{What is the dominant color in the image? The relevant options are B: brown
and C: black.}
{B (brown)}
{C receives 17 rollouts and B receives 15. Thus, the pool contains a substantial
minority of correct trajectories.}
{Self-Consistency & C & \qualincorrect & 17 votes \\
 LookBack top-1 & C & \qualincorrect & -- \\
 LookBack + RWC & C & \qualincorrect & 9.94 \\
 VisConf + RWC & C & \qualincorrect & 8.87 \\}
{This is a shared failure despite the presence of substantial correct evidence.
The C trajectories emphasize individually black objects, such as clothing,
instruments, and acoustic panels, while the B trajectories attend to the
image-wide brown or sepia cast. Both quality scores reinforce the narrow
incorrect majority rather than recovering the correct minority.}

\let\qualcase\relax
\let\qualcorrect\relax
\let\qualincorrect\relax
%%%%%%%%%%%%%%%%%%%%%%%%%%%%%%%%%%%%%%%%%%%%%%%%%%%%%%%%%%%%

% \newpage
% \input{checklist.tex}

\end{document}